\documentclass{article}

\usepackage{microtype}
\usepackage{graphicx}
\usepackage{subcaption}
\usepackage{booktabs} 
\usepackage{enumitem}

\usepackage{hyperref}

\usepackage[accepted]{style/icml2026}

\usepackage{amsmath}
\usepackage{amssymb}
\usepackage{mathtools}
\usepackage{amsthm}
\usepackage{amsmath}
\usepackage{bbm}
\usepackage{multirow}
\usepackage[table]{xcolor}
\usepackage{booktabs}
\usepackage{makecell}
\usepackage{wrapfig}

\DeclareMathOperator*{\argmax}{argmax}
\DeclareMathOperator*{\argmin}{argmin}

\definecolor{table_color}{HTML}{DAE8FC}
\definecolor{base_color}{HTML}{EFF7D7}
\definecolor{adapt_color}{HTML}{F7E8DA}

\newcommand{\RN}[1]{\uppercase\expandafter{\romannumeral#1}}

\usepackage[capitalize,noabbrev]{cleveref}
\usepackage{fontawesome5} 
\theoremstyle{plain}

\theoremstyle{definition}

\theoremstyle{remark}

\usepackage[textsize=tiny]{todonotes}

\icmltitlerunning{SuCo: Sufficiency-guided Continuous Adaptive Reasoning}

\begin{document}

\twocolumn[
  \icmltitle{SuCo: Sufficiency-guided Continuous Adaptive Reasoning}



  \icmlsetsymbol{equal}{*}

  \begin{icmlauthorlist}
    \icmlauthor{Jiahao Wang}{hitsz}
    \icmlauthor{Bingyu Liang}{hitsz}
    \icmlauthor{Chenhao Hu}{hitsz}
    \icmlauthor{Longhui Zhang \textsuperscript{\faIcon[regular]{envelope}}}{hitsz}
     \icmlauthor{Xuebo Liu}{hitsz}
    \icmlauthor{Min Zhang}{hitsz} \\
\icmlauthor{Jing Li \textsuperscript{\faIcon[regular]{envelope}}}{hitsz}
\icmlauthor{Xuelong Li \textsuperscript{\faIcon[regular]{envelope}}}{TeleAI}
  \end{icmlauthorlist}

  \icmlaffiliation{hitsz}{Harbin Institute of Technology (Shenzhen), Guangdong, China.}
  \icmlaffiliation{TeleAI}{TeleAI of China Telecom, China}

\icmlcorrespondingauthor{Longhui Zhang}{longhuizhang97@gmail.com}
\icmlcorrespondingauthor{Jing Li}{jingli.phd@hotmail.com}
\icmlcorrespondingauthor{Xuelong Li}{xuelong\_li@ieee.org.}

  \icmlkeywords{Machine Learning, ICML}

  \vskip 0.3in
]



\printAffiliationsAndNotice{}  

\newcommand{\ourmethod}{SuCo}

\begin{abstract}
Despite remarkable performance on complex tasks, Large Reasoning Models (LRMs) often generate excessively long Chain-of-Thoughts (CoT), inflating computational costs even for simple queries. 
Existing efforts to mitigate this inefficiency typically rely on discrete reasoning modes or fixed budget tiers, lacking a principled criterion of when reasoning is sufficient.
In this work, we introduce \emph{Minimal Sufficient CoT} (MSC), 
defined as the shortest prefix of a CoT trajectory which is adequate for producing the correct answer.
We empirically show that MSC not only reduces reasoning tokens, but also improves accuracy across difficulty levels.
Building on MSC, we propose \emph{Sufficiency-guided Continuous Adaptive Reasoning} (SuCo), 
a two-stage training framework for autonomous reasoning control along a continuous spectrum.
In stage \RN{1},
\emph{MSC-Aligned Fine-Tuning} (MFT) 
constructs MSC data using problem-adaptive sufficiency thresholds that naturally scale with question difficulty,
then fine-tunes the model to internalize concise yet sufficient reasoning patterns. 
In stage \RN{2}, 
\emph{Sufficiency-Aware Policy Optimization} (SAPO) 
further optimizes the model through reinforcement learning with dynamic complexity tracking and sufficiency-aware rewards that penalize both over- and under-thinking.
Extensive experiments across mathematics, code, and science benchmarks show that SuCo consistently achieves improvements in both accuracy and reasoning efficiency.
\end{abstract}
\section{Introduction}
Large Language Models (LLMs) have demonstrated impressive capabilities across a wide range of tasks~\citep{zhao2023survey, wang-etal-2025-system-report, zhang-etal-2025-speed},
yet continue to struggle with complex problems requiring multi-step reasoning~\citep{gsm8k}.
To address this limitation, recent work has introduced \emph{Large Reasoning Models} (LRMs), which explicitly generate intermediate reasoning steps via Chain-of-Thoughts (CoT)~\citep{wei2022chain}.
By performing step-by-step logical thinking before arriving at final answers,
LRMs such as DeepSeek-R1~\citep{deepseekr1} and OpenAI o1~\citep{openai} achieve substantial gains over standard LLMs on challenging benchmarks~\citep{hou2025advancing,xu2025towards}.

\begin{figure}[t]
  \begin{center}
    \centerline{\includegraphics[width=\columnwidth]{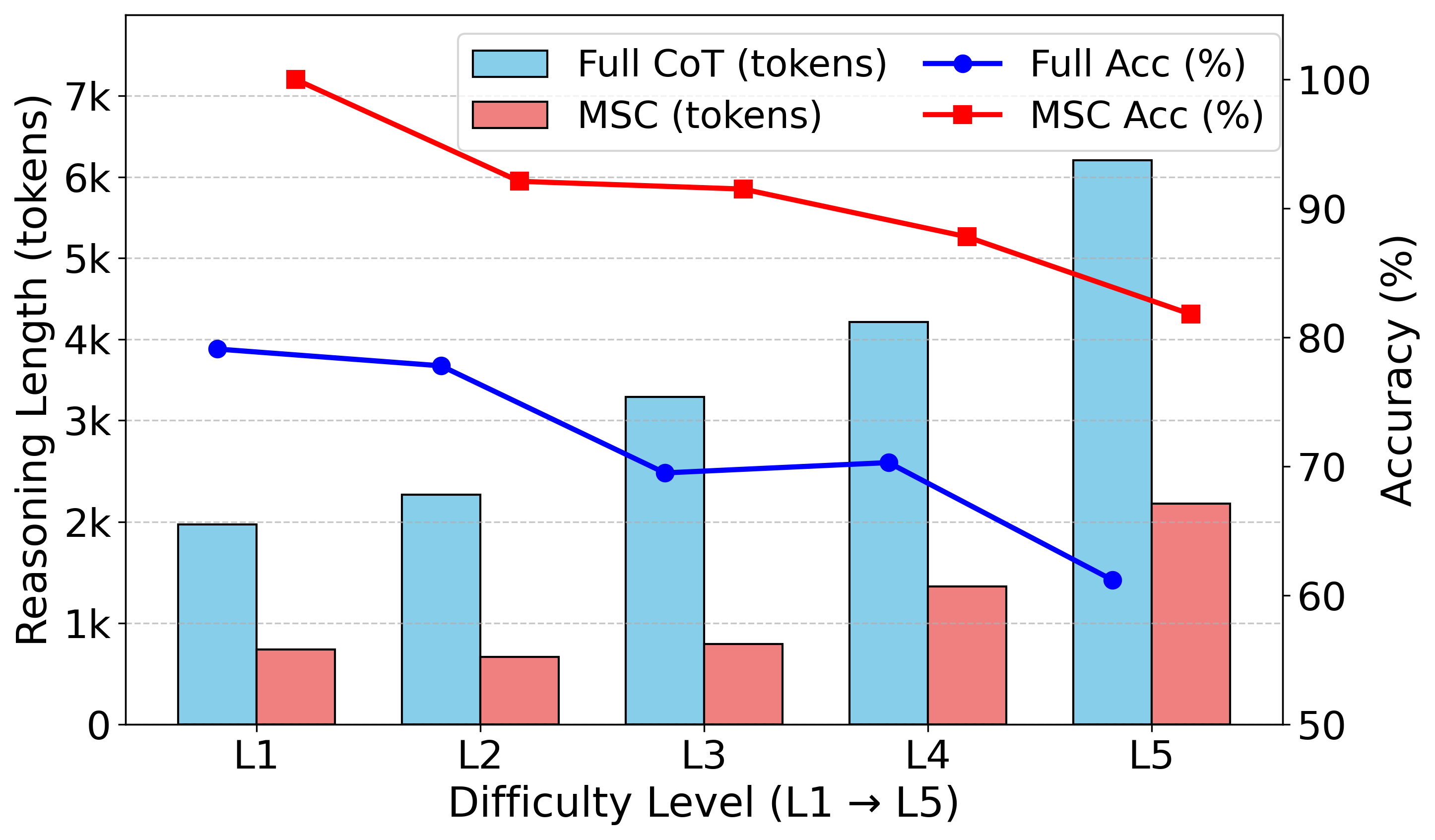}}
    \caption{
      MSC vs.\ Full CoT on Qwen3-8B across MATH difficulty levels. 
      \textbf{Left axis ($\downarrow$)}: reasoning tokens.
      \textbf{Right axis ($\uparrow$)}: accuracy. 
      At each difficulty level, MSC achieves higher accuracy with significantly fewer tokens.
    }
    \label{fig:msc}
  \end{center}
  \vspace{-3.0mm}
\end{figure}

Despite these advances, current LRMs suffer from \emph{redundant reasoning}~\citep{sui2025stop}. 
Even for simple queries, they tend to generate exhaustive reasoning chains, 
incurring substantial computational costs and inference latency~\citep{aggarwal2025l1}.
Such inefficiency limits practical deployment in real-time applications (e.g., online coding assistants~\citep{jimenez2023swe}) and resource-constrained environments (e.g., edge devices~\citep{zhang2024tinyllama}).

To mitigate redundancy, recent studies have developed \emph{Adaptive Large Reasoning Models} (ALRMs),
which aim to adjust reasoning effort according to problem complexity~\citep{sui2025stop, wu2025efficiency}.
These approaches can be broadly categorized into two paradigms.
\emph{User-controlled methods} require explicit prompts to select reasoning behaviors.
For example, Qwen3~\citep{qwen3} enables manual on/off switching, while GPT-OSS~\citep{gptoss} provides multiple predefined reasoning strategies.
In contrast,
\emph{model-driven methods} allow autonomous reasoning decisions.
AdaCoT~\citep{adacot} employs external assessors, whereas LHRM~\citep{lhrm} assigns reasoning status based on domain labels.
Despite their differences, existing ALRMs fundamentally rely on \textbf{discrete mode selection}.
Reasoning effort is adjusted by switching among a finite set of manually specified options,
rather than being calibrated in a continuous manner.

We posit that an ideal ALRM requires: 
(1) reasoning length scales with problem difficulty,
(2) autonomous resource allocation without intervention, and 
(3) optimal performance with minimal reasoning.
However, this raises a counterintuitive question:
According to the test-time scaling laws~\citep{snell2024scaling,brown2024large}, performance typically improves with more reasoning.
\textbf{Can models actually perform better with less reasoning?}  

We provide an affirmative answer by introducing \textbf{Minimal Sufficient CoT (MSC)} —
the shortest reasoning prefix of a CoT trajectory that is sufficient to yield the correct answer.
As illustrated in Figure~\ref{fig:msc}, 
across all five difficulty levels of the MATH benchmark~\citep{math}, 
MSC dramatically reduces reasoning tokens 
while consistently outperforming full CoT in accuracy.
This reveals that rather than blindly scaling reasoning resources, 
test-time adaptation offers a more efficient solution.

Building on this insight,
we propose \textbf{Su}fficiency-guided 
\textbf{Co}ntinuous Adaptive Reasoning (\textbf{SuCo}),
a two-stage training framework enabling continuous reasoning control.
Unlike prior discrete approaches that depend on external classifiers or predefined budget tiers, SuCo introduces problem-adaptive sufficiency thresholds that naturally adjust to question difficulty.
In Stage \RN{1},
\emph{MSC-Aligned Fine-Tuning} (MFT) constructs an MSC dataset from full CoT trajectories, then performs supervised fine-tuning (SFT) to internalize concise yet sufficient reasoning patterns.
In Stage \RN{2}, \emph{Sufficiency-Aware Policy Optimization} (SAPO) further trains the model to dynamically allocate reasoning effort through reinforcement learning (RL).
Critically, SAPO maintains a dynamic complexity pool to track evolving reasoning distributions during training,
and employs sufficiency-aware rewards that penalize both insufficient and excessive reasoning.

Extensive experiments are conducted across mathematics, code, and science domains at both 1.5B and 7B model scales.
Results demonstrate that SuCo achieves superior accuracy with substantially fewer reasoning tokens,
outperforming full CoT and ALRM baselines.

Our key contributions are summarized as follows:

\begin{itemize}[noitemsep,nolistsep]
    \item We formalize MSC, providing a principled sufficiency criterion revealing that models can achieve stronger performance with less reasoning.
    
    \item We propose SuCo, a two-stage training paradigm for continuous and autonomous reasoning control without discrete modes or external intervention.
    
    \item Comprehensive experiments spanning diverse domains demonstrate the effectiveness of our SuCo.
\end{itemize}
\section{Related Work}
\paragraph{Large Reasoning Models.}
Large Reasoning Models (LRMs) extend Large Language Models (LLMs) by explicitly generating intermediate reasoning steps via Chain-of-Thoughts (CoT),
which has been shown to substantially improve performance on challenging multi-step tasks~\citep{wei2022chain,kojima2022large}.
Building on this paradigm, 
recent LRMs such as DeepSeek-R1~\citep{deepseekr1}, OpenAI o1~\citep{openai}, and Qwen3~\citep{qwen3} further strengthen reasoning capabilities through large-scale supervised fine-tuning (SFT) on high-quality CoT data, often combined with reinforcement learning (RL) with curated rewards. 
Despite these advances, 
current LRMs frequently produce unnecessarily verbose reasoning even for trivial queries, 
incurring significant inference overhead and motivating the need for more efficient reasoning control.

\paragraph{Adaptive Large Reasoning Models.}
To mitigate reasoning redundancy, 
recent efforts have explored Adaptive Large Reasoning Models (ALRMs) that modulate reasoning length based on problem difficulty. 
Early approaches primarily focus on \emph{binary triggering} of reasoning.
AdaCoT~\citep{adacot} employs an external model to decide whether to activate CoT;
AdaptThink~\citep{adaptthink} formulates reasoning activation as a constrained optimization problem;
LHRMs~\citep{lhrm} assigns reasoning behaviors using coarse domain-level labels (e.g., math vs.\ chat). 
Beyond binary control, subsequent methods investigate \emph{multi-mode reasoning}. 
SABER~\citep{saber} and ThinkDial~\citep{thinkdial} introduce multiple predefined reasoning strategies or budget tiers, selected via system prompts.
Additional recent efforts have explored more fine-grained control.
 ThinkPrune~\citep{hou2025thinkprune} applies reinforcement learning to prune long reasoning chains,
while CyclicReflex~\citep{fan2026cyclicreflex} schedules reflection tokens cyclically to balance depth and efficiency.
AlphaOne~\citep{zhang2025alphaone} explores dual-speed reasoning at test time, enabling models to adaptively think slow or fast.
Complementary analyses have also highlighted the phenomena of \emph{underthinking}~\citep{wang2025thoughts} and the mirage of test-time scaling~\citep{ghosal2026does}, 
which further motivate principled control over reasoning effort.

Despite their progress, existing ALRMs share a fundamental limitation:
reasoning effort is regulated through \textbf{discrete specified modes},
supported by coarse supervision signals such as external estimators,
predefined data categories, or heuristic length constraints.
Such discrete control overlooks the internal logical sufficiency of reasoning trajectories
and lacks the flexibility to finely calibrate reasoning depth in a problem-specific manner.

In contrast, our work proposes \emph{continuous adaptive reasoning} grounded in the concept of
\emph{Minimal Sufficient CoT} (MSC).
By introducing a principled sufficiency criterion, we enable fine-grained assessment of whether a reasoning prefix is adequate to support a confident answer.
Unlike discrete modes or fixed truncation rules, 
our sufficiency-aware training empowers the model to autonomously calibrate its reasoning effort along a continuous spectrum.
\begin{figure*}[ht]  
\centering
\includegraphics[width=\textwidth]{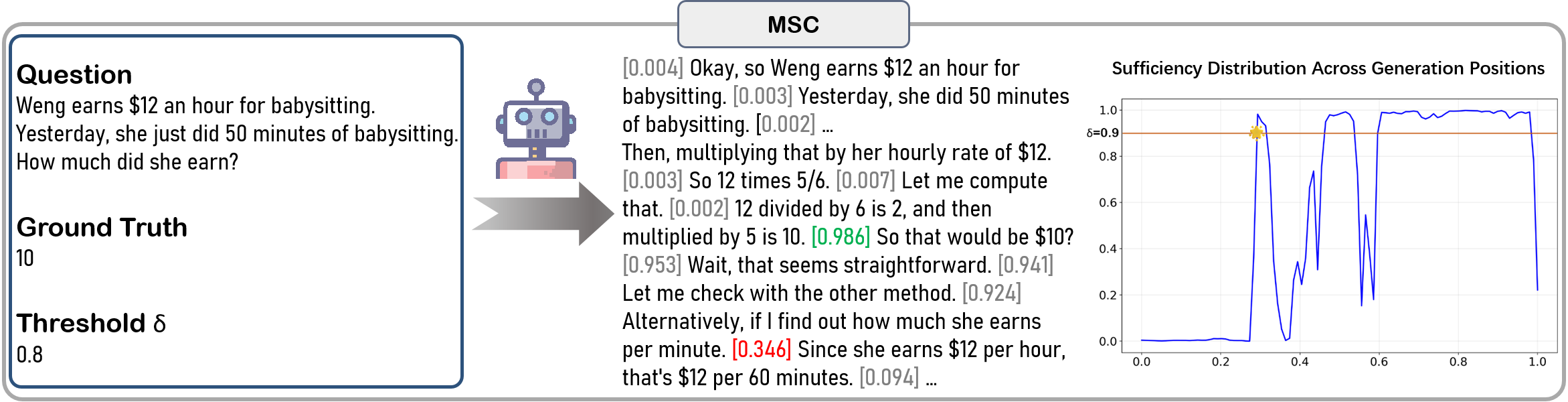}  
\caption{
\textbf{Illustration of Minimal Sufficient CoT (MSC).}
For a given question, sufficiency score (geometric mean over ground-truth answer tokens) is computed at each generation position.
The MSC is the shortest prefix exceeding the adaptive threshold $\delta$.
As shown, once the sufficiency threshold is reached, 
extended \emph{waiting} or self-verification steps lead to a rapid decline in sufficiency,
indicating that additional reasoning contributes little benefit and may even degrade confidence.
}
\label{fig:suco}
\end{figure*}

\section{Methodology}

\subsection{Problem Formulation}
\paragraph{Notation.}
Consider a dataset $\mathcal{D}$ of question-answer pairs $(x, y^*)$, 
where $x$ denotes an input question and $y^*$ the ground-truth answer. 
Given $x$, a reasoning model $\pi_{\theta}$ generates a CoT
trajectory $z = (z_1, z_2, \ldots, z_{L_z})$, 
with $L_z$ sentences and a total of $\|z\|$ tokens. 
Conditioned on $x$ and $z$, the model generates the final answer: 
$y \sim \pi_{\theta}(\cdot \mid x, z)$.

\subsection{MSC: Minimal Sufficient CoT}
Figure~\ref{fig:suco} provides an intuitive illustration of MSC.
\paragraph{Reasoning Sufficiency.}
To quantify how well a reasoning trajectory supports the ground-truth,
we define the \emph{reasoning sufficiency}:
\begin{equation}
\label{eq:sufficiency}
\mathcal{S}_{\theta}(z \mid x, y^*) := \left( \prod_{i=1}^{\|y^*\|} \pi_\theta(y^*_i \mid x, z, y^*_{<i}) \right)^{1/\|y^*\|}
\end{equation}
The most natural signal is the joint probability $\prod_{i} \pi_\theta(y^*_i \mid x, z, y^*_{<i})$. 
However, it decays exponentially with answer length, making it fragile for long sequences.
To address this, we employ the geometric mean, which normalizes the joint probability into a per-token average.
We empirically validate this choice in Appendix~\ref{app:sufficiency_ablation}.

\paragraph{Sufficient CoT.}
Then we can determine whether reasoning is adequate 
by introducing a confidence threshold $\delta \in [0, 1]$. 
A trajectory $z$ is termed \emph{$\delta$-sufficient} if
$\mathcal{S}_{\theta}(z|x, y^*) \ge \delta$.

\paragraph{MSC Definition.}
We further define the MSC as the shortest reasoning prefix satisfying sufficiency.
We identify MSC at the sentence level, 
as sentence boundaries naturally correspond to atomic reasoning steps, 
and avoid fragmentary truncation that may distort logical structure.
We  say the prefix $z_{<t^*}$ is a $\delta$-MSC if and only if:
\begin{equation}
\begin{cases}
\mathcal{S}_\theta(z_{<t^*} \mid x, y^*) \geq \delta & \text{(Sufficiency)} \\
\mathcal{S}_\theta(z_{<t} \mid x, y^*) < \delta, \quad \forall t < t^* & \text{(Minimality)}
\end{cases}
\end{equation}

\paragraph{Problem-Adaptive Threshold.}
A fixed threshold $\delta$ applies the same confidence bar uniformly across all problems,
regardless of their inherent difficulty.
However, for simple problems, a high $\delta$ retains unnecessary reasoning,
while for hard problems, a low $\delta$ may truncate critical reasoning steps prematurely.
We therefore introduce a \textit{problem-adaptive threshold}:
\begin{equation}
\label{eq:threshold}
\delta(x) = \delta_0 + \alpha \cdot \mathcal{C}(x)
\end{equation}
where $\delta_0$ is the base value, 
$\alpha$ controls sensitivity to complexity,
and $\mathcal{C}(x) \in [0,1]$ denotes problem complexity.
This produces a more discriminative MSC distribution across difficulty levels,
providing a stronger adaptive prior for subsequent training.

\paragraph{Percentile-Based Complexity Estimation.}
We estimate complexity as the percentile rank of its reasoning length in the dataset:
reasoning length serves as a practical proxy for problem complexity, as empirically supported in Figure~\ref{fig:msc}.
Formally, given a dataset $\mathcal{D} = \{(x_i, y_i^*, z_i)\}_{i=1}^N$, 
we define:
\begin{equation}
\label{eq:complexity}
\mathcal{C}(x_i) = \frac{1}{N} \sum_{j=1}^N \mathbbm{1}[\|z_j\| \leq \|z_i\|]
\end{equation}
This percentile-based measure is robust to outliers 
and yields values uniformly distributed in $[0,1]$,
ensuring stable threshold scaling across problems.

\subsection{Stage \RN{1}: MSC-Aligned Fine-Tuning}
The first stage, termed MSC-Aligned Fine-Tuning (MFT), aligns the model to produce concise yet sufficient reasoning 
through SFT on a curated MSC dataset. 
This stage consists of two steps: 
(1) constructing MSC data from full CoT trajectories, and 
(2) fine-tuning the model to internalize adaptive reasoning patterns.

\paragraph{MSC Data Construction.}
From source dataset $\mathcal{D}_{\text{src}} = \{(x_i, y_i^*)\}_{i=1}^N$,
a strong reasoning model $\mathcal{M}_{\text{LRM}}$ generates full CoT  and answers: 
$(\hat{z}_i, \hat{y}_i) \sim \mathcal{M}_{\text{LRM}}(x_i)$.
We then extract MSC from each trajectory via the following procedure:

{\normalsize $\blacktriangleright$ } \textit{(i) Compute adaptive thresholds.}
With access to all trajectory lengths $\{\|\hat{z}_i\|\}_{i=1}^N$,
we derive per-sample complexity $\mathcal{C}(x_i)$ and threshold $\delta(x_i)$ using Eq.~\ref{eq:complexity} and Eq.~\ref{eq:threshold}.

\begin{algorithm}[t]
\small
\caption{MSC Dataset Construction}
\label{alg:msc}
\begin{algorithmic}[1]
\REQUIRE Source dataset $\mathcal{D}_{\text{src}}$; 
         models $\mathcal{M}_{\text{LRM}}$, 
         $\mathcal{M}_{\text{refine}}$,
         $\pi_\theta$;
         hyperparameters $\delta_0$, $\alpha$, $L_{\min}$

\FOR{each $(x_i, y_i^*) \in \mathcal{D}_{\text{src}}$}
    \STATE $(\hat{z}_i, \hat{y}_i) \sim \mathcal{M}_{\text{LRM}}(x_i)$
\ENDFOR
\STATE $\mathcal{D}_{\text{full}} \gets \{(x_i, y_i^*, \hat{z}_i, \hat{y}_i)\}_{i=1}^N$

\FOR{each $i \in [1, N]$}
    \STATE $\mathcal{C}(x_i) \gets \frac{1}{N} \sum_{j=1}^N \mathbbm{1}[\|\hat{z}_j\| \leq \|\hat{z}_i\|]$ 
    \STATE $\delta(x_i) \gets \delta_0 + \alpha \cdot \mathcal{C}(x_i)$
    
    \STATE $t^* \gets \argmin_{t \in [0, L_{\hat{z}_i}]} \mathcal{S}_\theta(\hat{z}_{i,<t} \mid x_i, y_i^*) \geq \delta(x_i)$
    \IF{no such $t$ exists}
        \STATE $t_i^* \gets \argmax_{t \in [0, L_{\hat{z}_i}]} \mathcal{S}_\theta(\hat{z}_{i,<t} \mid x_i, y_i^*)$
    \ENDIF
    \STATE $z_i^{\text{raw}} \gets \hat{z}_{i,<t_i^*}$
    \IF{$L_{z_i^{\text{raw}}} \leq L_{\min}$}
        \STATE $z_i^{\text{MSC}} \gets \varnothing$
    \ELSE
        \STATE $z_i^{\text{MSC}} \gets \mathcal{M}_{\text{refine}}(x_i, z_i^{\text{raw}}, \hat{y}_i)$
    \ENDIF
\ENDFOR

\STATE \textbf{return} $\mathcal{D}_{\text{MSC}} = \{(x_i, z_i^{\text{MSC}}, \hat{y}_i)\}_{i=1}^N$
\end{algorithmic}
\end{algorithm}

{\normalsize $\blacktriangleright$ } \textit{(ii) Identify raw MSC prefixes.} 
For each sample, we scan sentence-level prefixes to find the minimal sufficient one:
\begin{equation}
\label{eq:msc_index}
t_i^* = \argmin_{t \in [0, L_{\hat{z}_i}]} \mathcal{S}_\theta(\hat{z}_{i,<t} \mid x_i, y_i^*) \geq \delta(x_i)
\end{equation}
If no prefix reaches $\delta(x_i)$, we select the most sufficient one:
\begin{equation}
t_i^* = \argmax_{t \in [0, L_{\hat{z}_i}]} \mathcal{S}_\theta(\hat{z}_{i,<t} \mid x_i, y_i^*).
\end{equation}
This yields a raw candidate: $z_i^{\text{raw}} = \hat{z}_{i,<t_i^*}$.

To avoid trivial fragments, 
we set $z_i^{\text{raw}}$ with empty string 
if $\|z_i^{\text{raw}}\| \leq L_{\min}$,
indicating that the question requires no explicit reasoning.

{\normalsize $\blacktriangleright$ } \textit{(iii) Refine MSC for coherence.} 
Raw truncation may leave logical gaps.
We use $\mathcal{M}_{\text{refine}}$
to polish each nonempty MSC with the following objectives:
(1) naturally derive the answer,
(2) eliminate redundancy, and
(3) preserve stylistic consistency.
This produces the final refined $z_i^{\text{MSC}}$.  

The final dataset is formatted as: 
\begin{equation}
\mathcal{D}_{\text{MSC}} = \left\{ \left( x_i, \texttt{<think>}\, z_i^{\text{MSC}}\, \texttt{</think>}\, \hat{y}_i \right) \right\}_{i=1}^N
\end{equation}
where $z_i^{\text{MSC}}$ can be empty for questions requiring no reasoning.
The complete procedure is detailed in Algorithm~\ref{alg:msc}.

\paragraph{Supervised Fine-Tuning.}
We fine-tune the base model by minimizing the negative log-likelihood over $\mathcal{D}_{\text{MSC}}$:
\begin{equation}
\begin{aligned}
\mathcal{L}_{\text{MFT}}(\theta) &= -\mathbb{E}_{(x_i,z_i^{\text{MSC}},\hat{y}_i)\sim\mathcal{D}_{\text{MSC}}} \\
\Bigg[ 
&\quad \log \pi_\theta(z_i^{\text{MSC}} \mid x_i) 
+ \log \pi_\theta(\hat{y}_i \mid x_i, z_i^{\text{MSC}}) 
\Bigg]
\end{aligned}
\end{equation}


\subsection{Stage \RN{2}: Sufficiency-Aware Policy Optimization}
\label{sec:sapo}
In the second stage, named Sufficiency-Aware Policy Optimization (SAPO),
we train the model to allocate reasoning steps during inference 
through RL with a dynamic complexity pool  and sufficiency-aware rewards.
We build upon Group Relative Policy Optimization (GRPO)~\citep{deepseekmath},
which samples multiple trajectories per question to enable
robust group-wise advantage estimation.

\paragraph{Dynamic Complexity Pool.}
A critical challenge in integrating MSC into online $\text{RL}$ is
that the reasoning length distribution shifts as the policy evolves.
The offline complexity estimates from MFT stage become obsolete.
Recomputing them over the entire dataset after each gradient step is computationally prohibitive.

Instead, we maintain an online \textit{dynamic complexity pool} $\mathcal{P} = \{\|z_i^{\text{avg}}\|\}_{i=1}^N$ 
that tracks the evolving reasoning length for each question $x_i$.
The pool is initialized from $\pi_{\text{MFT}}$ on the RL training data, i.e., $\|z_i^{\text{avg}}\| \gets \mathbb{E}_{z \sim \pi_{\text{MFT}}(\cdot|x_i)}[\|z\|]$.
For each training batch, 
we update the pool via exponential moving average (EMA):
\begin{equation}
\label{eq:ema_update}
\|z_i^{\text{avg}}\| \gets (1-\eta)\cdot \|z_i^{\text{avg}}\| + \eta \cdot \frac{1}{K}\sum_{k=1}^K \|z_i^{(k)}\|,
\end{equation}
where $\eta \in [0,1]$ controls the update rate, 
and $\{z_i^{(k)}\}_{k=1}^K$ are the $K$ rollout trajectories for $x_i$ in the current batch.

From $\mathcal{P}$,
we recompute complexity scores $\mathcal{C}(x_i)$ and thresholds $\delta(x_i)$ 
via Eq.~\ref{eq:complexity} and Eq.~\ref{eq:threshold}.  
This mechanism ensures sufficiency targets aligned with the policy’s current behavior,
providing stable reward signals at negligible extra cost.

\paragraph{Sufficiency-Aware Reward Shaping.}
The total reward:
\begin{equation}
\label{eq:total_reward}
\mathcal{R}(z, y \mid x, y^*) = \mathcal{R}_{\text{cor}}(y) + \mathcal{R}_{\text{format}}(z,y) + \beta \cdot \mathcal{R}_{\text{suff}}(z \mid x, y^*)
\end{equation}
where $\mathcal{R}_{\text{cor}}$ rewards correct answers,
and $\mathcal{R}_{\text{format}}$ ensures proper use of \texttt{<think>...</think>} delimiters.

The sufficiency reward $R_{\text{suff}}$ uses 
the current adaptive threshold $\delta(x)$ from the dynamic pool. 
For each trajectory $z$,
we identify the earliest sufficient prefix $z_{<t_i^*}$ using Eq.~\ref{eq:msc_index}. 
If no prefix satisfies the threshold, we set $t^*=\infty$.
The reward penalizes both over-thinking and under-thinking:
\begin{equation}
\label{eq:suff_reward}
\begin{aligned}
\mathcal{R}_{\text{suff}}(x, z, y) &= \;
\underbrace{ -\lambda_{over} \cdot \mathbbm{1}[L_z > t^* + \epsilon] }_{\text{over-thinking}} \\[4pt]
&\quad - \underbrace{ \mathbbm{1}[y \neq y^*] \cdot \lambda_{under} \cdot \mathbbm{1}[L_z < t^*] }_{\text{under-thinking}}
\end{aligned}
\end{equation}
The tolerance $\epsilon$ allows minor deviations beyond sufficiency,
and the under-thinking penalty applies only to incorrect generations.

\section{Experiments}

\subsection{ Experimental Settings}

\paragraph{Training datasets.} We train SuCo on reasoning datasets spanning mathematics, code, and science. The data are drawn from five sources:  Llama-Nemotron Post-Training Dataset~\citep{llama},  Mixture-of-Thoughts~\citep{mot},  OpenR1-Math-220k~\citep{openr1math},  OpenCodeReasoning~\citep{opencodereasoning},  and s1K-1.1~\citep{s1k}. These datasets contain reasoning chains distilled from state-of-the-art LRMs. After filtering and deduplication, we construct the corresponding MSC for each sample following Algorithm~\ref{alg:msc}. We further remove low-quality MSC samples using LLM-based quality assessment. Both MSC refinement and quality assessment are performed  with Qwen3-Next-80B-A3B-Instruct~\citep{qwen3}. This process yields 270,011 high-quality training samples. Detailed construction procedures and statistics are provided in Appendix~\ref{sec:dataset_construction}. Figure~\ref{fig:dataset_token} compares the token length distributions of full CoT and MSC across the training corpus. We use the full MSC dataset for Stage~\RN{1}, and sample a subset of the data for RL in Stage~\RN{2}.

\begin{figure}[ht]
\vskip 0.2in
\begin{center}
\centerline{\includegraphics[width=\columnwidth]{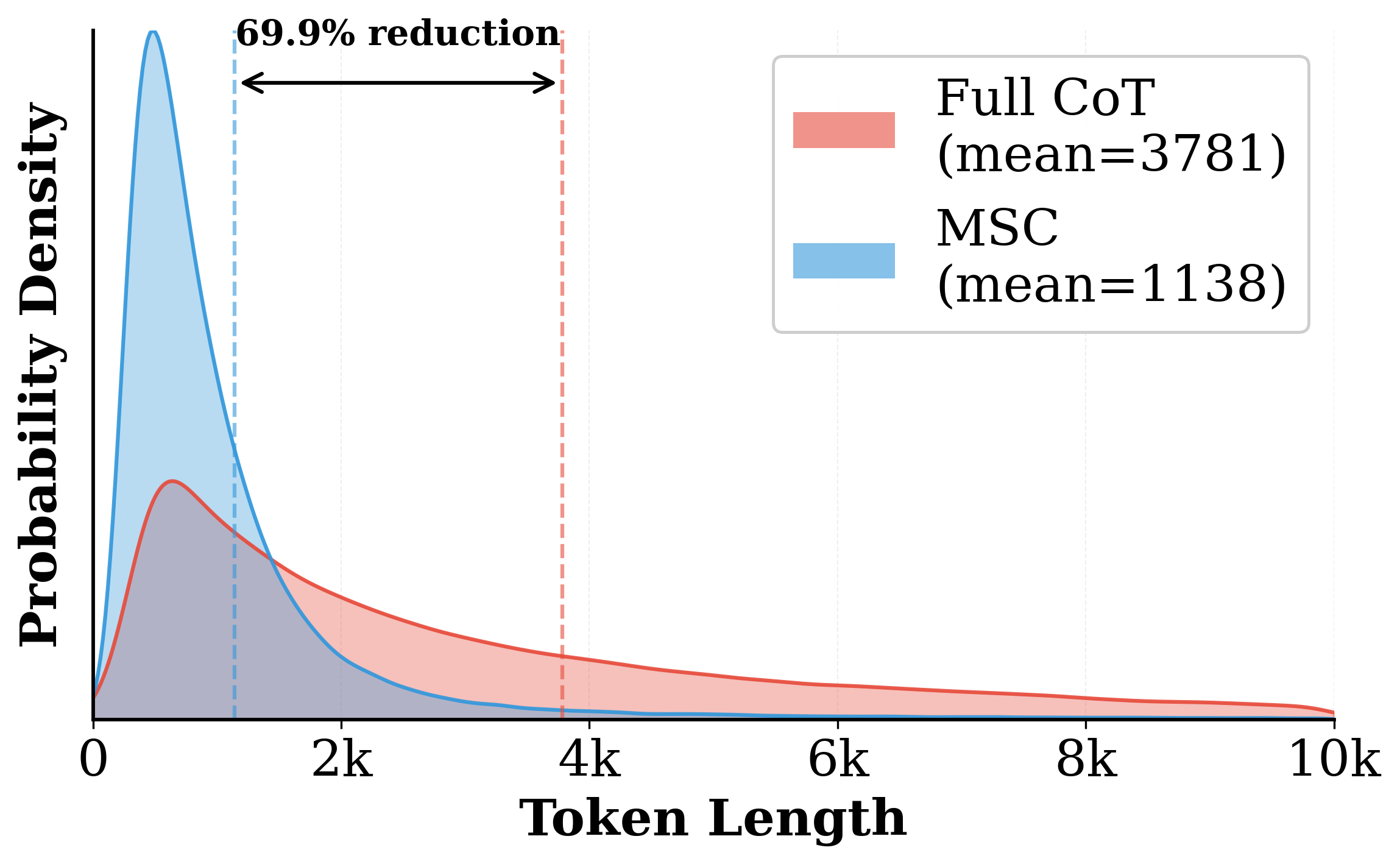}}
\caption{Token length distribution comparison between full CoT and MSC across training datasets.}
\label{fig:dataset_token}
\end{center}
\vskip -0.2in
\end{figure}

\paragraph{Implementation details.}
All trainings are performed on 8 $\times$ NVIDIA H100 80GB GPUs.
\textbf{MFT Stage.}
We set the base threshold $\delta_0=0.5$ and the sensitivity coefficient $\alpha=0.4$, resulting in problem-adaptive thresholds $\delta(x) \in [0.5, 0.9]$.
The minimum reasoning length is fixed to $L_{\min}=5$ sentences to filter trivial fragments.
We train for 3 epochs with a learning rate of $1\times10^{-4}$.
\textbf{SAPO Stage.}
The dynamic complexity pool is initialized using predictions from the MFT model and updated during training with an EMA rate $\eta=0.1$.
For each training instance, we sample $K=8$ rollout trajectories.
The sufficiency reward weight is set to $\beta=1.0$, with over- and under-thinking penalties $\lambda_{\text{over}}=\lambda_{\text{under}}=0.5$ and a tolerance margin $\epsilon=2$ sentences.
We train using Group Relative Policy Optimization (GRPO)~\citep{deepseekmath} with learning rate $1 \times 10^{-6}$, a batch size of 128 and a micro batch size of 8.

\leavevmode\pdfdest name{dummy1} xyz
\begin{table*}[t]
\centering
\small
\renewcommand{\arraystretch}{0.92}
\caption{Main results on mathematics (GSM8K, MATH-500, AMC23, AIME25), code (MBPP, LiveCodeBench-V6), and science (MMLU-STEM, GPQA-Diamond) benchmarks.
Best results in each section are \textbf{bolded}, second best are \underline{underlined}.}
\label{tab:main_results}
\begin{tabular}{l|cccc|cc|cc|c}
\toprule
& \multicolumn{4}{c|}{\textbf{Math}}
& \multicolumn{2}{c|}{\textbf{Code}}
& \multicolumn{2}{c|}{\textbf{Science}}
& \multirow{2}{*}{\textbf{Avg.}} \\
\textbf{Methods}
& \textbf{GSM8K}
& \textbf{MATH500}
& \textbf{AMC23}
& \textbf{AIME25}
& \textbf{MBPP}
& \textbf{Live-V6}
& \textbf{MMLU-S}
& \textbf{GPQA-D}
& \\
\midrule
\midrule
\multicolumn{10}{c}{\textbf{(\RN{1}) Reasoning Correctness Evaluation: Response Accuracy (\%) $\uparrow$}} \\
\midrule
\midrule
\multicolumn{10}{l}{\emph{Qwen2.5-1.5B}} \\
\rowcolor{base_color}
Math-Base & 40.1 & 22.6 & 23.9 & 3.3 & 4.0 & 0.6 & 14.5 & 4.0 & 14.1 \\
\rowcolor{base_color}
Math-Instruct & 79.0 & 72.4 & 43.5 & 6.7 & 6.1 & 2.3 & 30.2 & 22.2 & 32.8 \\
\rowcolor{base_color}
DeepSeek-R1-Distill & 80.3 & 80.6 & 56.5 & 26.7 & 41.0 & 17.1 & 33.8 & 25.3 & 45.2 \\
\rowcolor{adapt_color}
AdaCoT & 82.7 & 83.2 & 62.5 & 27.3 & \underline{44.3} & 17.7 & 33.9 & 26.3 & 47.2 \\
\rowcolor{adapt_color}
AdaptThink & 83.2 & 83.8 & 65.8 & 28.3 & 42.1 & 18.3 & 34.4 & 26.8 & 47.8 \\
\rowcolor{adapt_color}
S-GRPO & 83.4 & 84.2 & 69.5 & 31.0 & 42.9 & 19.4 & 34.8 & 28.3 & 49.2 \\
\rowcolor{adapt_color}
LHRMs & \underline{85.7} & \underline{84.4} & \underline{70.0} & \textbf{34.0} & 43.1 & \underline{20.6} & \underline{35.7} & \underline{30.3} & \underline{50.5} \\
\rowcolor{table_color}
\textbf{SuCo (Ours)} & \textbf{87.7} & \textbf{86.8} & \textbf{73.8} & \underline{33.7} & \textbf{48.5} & \textbf{22.3} & \textbf{38.6} & \textbf{33.3} & \textbf{53.1} \\
\midrule
\multicolumn{10}{l}{\emph{Qwen2.5-7B}} \\
\rowcolor{base_color}
Math-Base & 61.8 & 54.2 & 36.2 & 7.0 & 13.1 & 1.1 & 26.4 & 13.1 & 26.6 \\
\rowcolor{base_color}
Math-Instruct & 87.0 & 72.4 & 53.5 & 12.3 & 26.4 & 8.0 & 51.3 & 32.3 & 42.9 \\
\rowcolor{base_color}
DeepSeek-R1-Distill & 89.3 & 89.0 & 75.5 & 49.7 & 57.6 & 31.4 & 67.6 & 45.5 & 63.2 \\
\rowcolor{adapt_color}
AdaCoT & 91.4 & 91.8 & 81.5 & 55.0 & 61.7 & 32.0 & 69.0 & 47.0 & 66.2 \\
\rowcolor{adapt_color}
AdaptThink & \underline{92.9} & 92.8 & 82.0 & 54.3 & 62.0 & 32.6 & 68.8 & 47.0 & 66.6 \\
\rowcolor{adapt_color}
S-GRPO & 92.8 & 92.2 & \textbf{90.5} & \underline{58.3} & 62.3 & 
\underline{35.4} & \underline{72.2} & \underline{51.5} & \underline{69.4} \\
\rowcolor{adapt_color}
LHRMs & 92.4 & \underline{93.0} & 87.3 & 57.7 & \underline{62.4} & \underline{35.4} & 71.4 & 49.5 & 68.6 \\
\rowcolor{table_color}
\textbf{SuCo (Ours)} & \textbf{93.9} & \textbf{93.6} & \underline{90.3} & \textbf{61.7} & \textbf{65.7} & \textbf{38.9} & \textbf{75.8} & \textbf{56.6} & \textbf{72.1} \\
\midrule
\midrule
\multicolumn{10}{c}{\textbf{(\RN{2}) Reasoning Efficiency Evaluation: Response Length (Tokens) $\downarrow$}} \\
\midrule
\midrule
\multicolumn{10}{l}{\emph{Qwen2.5-1.5B}} \\
\rowcolor{base_color}
DeepSeek-R1-Distill & 501 & 4,260 & 6,768 & 11,239 & 3,511 & 11,073 & 1,956 & 6,582 & 5,736 \\
\rowcolor{adapt_color}
AdaCoT & 443 & 1,479 & 2,936 & 6,271 & 1,720 & 6,455 & 1,029 & 3,279 & 2,952 \\
\rowcolor{adapt_color}
AdaptThink & 337 & 1,564 & 2,740 & 6,513 & 1,422 & 6,689 & 995 & 2,914 & 2,897 \\
\rowcolor{adapt_color}
S-GRPO & \underline{297} & 1,377 & 3,081 & 6,640 & 1,481 & 5,293 & 812 & 2,828 & 2,726 \\
\rowcolor{adapt_color}
LHRMs & \textbf{242} & \underline{1,252} & \underline{2,477} & \textbf{3,257} & \underline{1,158} & \underline{4,716} & \underline{955} & \underline{2,381} & \underline{2,055} \\
\rowcolor{table_color}
\textbf{SuCo (Ours)} & 304 & \textbf{538} & \textbf{1,687} & \underline{3,484} & \textbf{930} & \textbf{2,629} & \textbf{745} & \textbf{1,550} & \textbf{1,483} \\
\midrule
\multicolumn{10}{l}{\emph{Qwen2.5-7B}} \\
\rowcolor{base_color}
DeepSeek-R1-Distill & 465 & 3,126 & 5,466 & 10,833 & 3,182 & 10,407 & 1,572 & 6,858 & 5,239 \\
\rowcolor{adapt_color}
AdaCoT & 422 & 1,387 & 3,115 & 7,648 & 1,423 & 8,153 & 1,214 & 3,993 & 3,419 \\
\rowcolor{adapt_color}
AdaptThink & \underline{247} & 1,424 & 2,834 & 7,842 & \underline{1,242} & 8,911 & 1,023 & 3,677 & 3,400 \\
\rowcolor{adapt_color}
S-GRPO & 267 & 988 & 2,247 & 5,259 & 1,356 & 6,534 & 717 & 2,453 & 2,478 \\
\rowcolor{adapt_color}
LHRMs & 274 & \underline{658} & \underline{1,525} & \underline{4,217} & 1,487 & \underline{4,294} & \underline{628} & \underline{2,042} & \underline{1,891} \\
\rowcolor{table_color}
\textbf{SuCo (Ours)} & \textbf{243} & \textbf{429} & \textbf{935} & \textbf{2,679} & \textbf{1,149} & \textbf{2,809} & \textbf{505} & \textbf{1,389} & \textbf{1,267} \\
\bottomrule
\end{tabular}
\end{table*}
\leavevmode\pdfdest name{dummy2} xyz
\paragraph{Benchmarks and Metrics.}
We conduct comprehensive evaluations across mathematics, code, and science domains, 
covering a broad range of problem difficulties.
For \textbf{mathematics}, we evaluate on 
GSM8K~\citep{gsm8k}, MATH-500~\citep{math500}, AMC 2023, and AIME 2025.
Due to the limited size of AMC 2023 (40 problems) and AIME 2025 (30 problems), 
each evaluation is repeated 10 times and results are averaged to reduce variance and improve statistical reliability.
For \textbf{code}, we use 
MBPP~\citep{mbpp}, and LiveCodeBench v6~\citep{livecodebench}.
For \textbf{science}, we test on MMLU-STEM~\citep{mmlu} and GPQA-Diamond~\citep{gpqa}.
Across all benchmarks, we report both accuracy and response length.

\paragraph{Baselines.}
We implement SuCo on Qwen2.5-Math-1.5B/7B-Base~\citep{qwen2.5_math} and compare against the following baselines at matched model scales.
\textbf{Standard Models.} 
We evaluate Qwen2.5-Math-Base, Qwen2.5-Math-Instruct, along with DeepSeek-R1-Distill-Qwen~\citep{deepseekr1} as the full CoT reasoning baseline.
\textbf{Adaptive Large Reasoning Models (ALRMs).}
We compare with four representative ALRMs: 
(1) AdaCoT~\citep{adacot} employs an external complexity assessor and PPO with Pareto optimization.
(2) AdaptThink~\citep{adaptthink} uses constrained RL for binary mode selection.
(3) S-GRPO~\citep{sgrpo} samples multiple early-exit positions with decaying rewards during RL training.
(4) LHRMs~\citep{lhrm} performs hybrid fine-tuning on categorized data followed by group policy optimization.
For fair comparison, AdaCoT and LHRMs are initialized from Qwen2.5-Math-Base and trained on the same source data as SuCo, while AdaptThink and S-GRPO follow their original implementations using DeepSeek-R1-Distill-Qwen as the base model.

\begin{table*}[t]
\centering
\small
\caption{Ablation results of MFT components. The results evaluate the effectiveness of MFT against the base model and full CoT training. We further analyze the sensitivity of sufficiency thresholds, complexity estimation strategies, and the impact of MSC refinement.}
\label{tab:mft_ablation}
\begin{tabular}{l l cc cc cc cc}
\toprule
\multirow{2}{*}{\textbf{Component}} & \multirow{2}{*}{\textbf{Method}} 
& \multicolumn{2}{c}{\textbf{Math}} 
& \multicolumn{2}{c}{\textbf{Code}} 
& \multicolumn{2}{c}{\textbf{Science}} 
& \multicolumn{2}{c}{\textbf{Avg.}} \\
\cmidrule(lr){3-4} \cmidrule(lr){5-6} \cmidrule(lr){7-8} \cmidrule(lr){9-10}
& & \textbf{Acc $\uparrow$} & \textbf{Tokens $\downarrow$} 
& \textbf{Acc $\uparrow$} & \textbf{Tokens $\downarrow$} 
& \textbf{Acc $\uparrow$} & \textbf{Tokens $\downarrow$} 
& \textbf{Acc $\uparrow$} & \textbf{Tokens $\downarrow$} \\
\midrule
\multirow{3}{*}{Overall} 
& Base       &  22.5 & - & 2.3 & - & 9.3  & - & 11.4 & -  \\
& Full       &  59.1 & 5,380 & 28.5 & 6,345 & 27.7 & 3,522 & 38.4 & 5,082  \\
\rowcolor{table_color}
& \textbf{MFT}  & 69.1 & 1,359 & 33.2 & 1,706 & 34.8 & 966 & 45.7 & 1,344  \\
\midrule
\multirow{5}{*}{Threshold}
& $\delta=0.9$ & 66.8 & 1,545 & 30.6 & 2,128 & 32.0 & 1,308 & 43.1 & 1,660 \\
& $\delta=0.8$ & 66.1 & 1,420 & 30.1 & 1,853 & 31.0 & 1,072 & 42.4 & 1,448 \\
& $\delta=0.7$ & 67.2 & 1,246 & 31.0 & 1,769 & 33.8 & 1,024 & 44.0 & 1,346 \\
& $\delta=0.6$ & 60.9 & 1,072 & 26.6 & 1,645 & 27.5 & 910 & 38.3 & 1,209 \\
& $\delta=0.5$ & 61.9 & 889 & 25.7 & 1,305 & 25.8 & 771 & 37.8 & 988 \\ 
\midrule
\multirow{3}{*}{Complexity}
& Min--Max    & 61.7 & 1,130 & 26.1 & 1,506 & 27.9 & 843 & 38.6 & 1,160  \\
& Log-Scaled  & 68.4 & 1,506 & 31.5 & 1,642 & 33.5 & 1,049 & 44.5 & 1,399  \\
\midrule
Refinement
& w/o refine & 65.9 & 2,220 & 31.8 & 2,741 & 32.7 & 1,607 & 43.5 & 2,189  \\
\bottomrule
\end{tabular}
\end{table*}

\subsection{Main Results}
\paragraph{Reasoning Correctness Evaluation.}
As shown in Table~\ref{tab:main_results} (\RN{1}),
across all model scales and domains,
SuCo consistently achieves the highest or near-highest accuracy.
At the 1.5B scale, SuCo attains an accuracy of 53.1\%, 
achieving a relative improvement of 5.1\% over the strongest adaptive baseline LHRMs
and 17.5\% over DeepSeek-R1-Distill-Qwen.
At the 7B scale, SuCo further improves to 72.1\% accuracy, exceeding  LHRMs by 5.1\% and DeepSeek-R1-Distill-Qwen by 14.1\%.
Notably, SuCo exhibits particularly strong gains on challenging benchmarks.
For example, on AIME25, SuCo attains 33.7\% accuracy at 1.5B scale and 61.7\% at 7B scale, 
corresponding to relative improvements of 26.2\% and 24.1\% 
over DeepSeek-R1-Distill-Qwen, respectively.

\paragraph{Reasoning Efficiency Evaluation.}
Table~\ref{tab:main_results} (\RN{2}) reports reasoning efficiency measured by average response length.
In addition to attaining higher accuracy, SuCo significantly reduces token consumption
compared to DeepSeek-R1-Distill-Qwen.
Across all benchmarks, SuCo reduces average response tokens by 74.1\% at the 1.5B scale
and by 75.8\% at the 7B scale, yielding substantial inference cost savings.
On AIME25 at 7B scale, SuCo achieves 24.1\% higher accuracy while using 75.3\% fewer tokens.
SuCo also outperforms other adaptive reasoning methods, 
confirming that sufficiency-aware training eliminates redundant reasoning without sacrificing decision quality.

\subsection{Ablation Study}
We analyze the contribution of each component in SuCo on Qwen2.5-Math-1.5B.
Additional ablation studies on hyperparameters ($L_{\min}$, $\epsilon$, $\eta$) are provided in Appendix~\ref{app:additional_ablations}.

\paragraph{MFT Ablations.}

\begin{figure}[ht]
\vskip 0.2in
\begin{center}
\centerline{\includegraphics[width=\columnwidth]{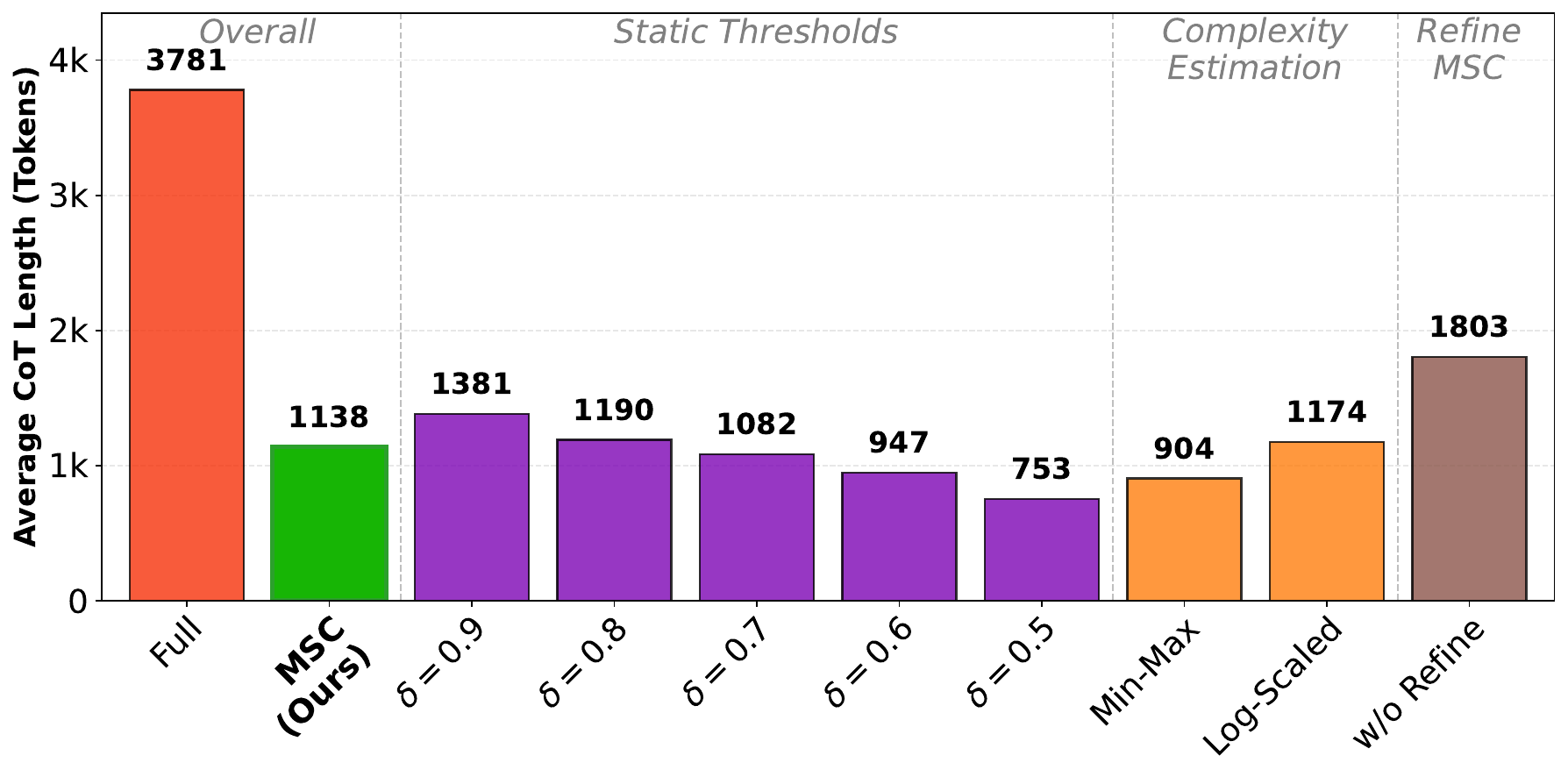}}
\caption{Distribution of reasoning lengths in training data constructed by different MSC variants.}
\label{fig:cot_length_distribution}
\end{center}
\vskip -0.2in
\end{figure}

Results are summarized in Table~\ref{tab:mft_ablation}, 
with training CoT length distributions illustrated in Figure~\ref{fig:cot_length_distribution}.

{\normalsize $\blacktriangleright$ } \textit{Overall Effectiveness.}
While full CoT training improves the base model from 11.4\% to 38.4\% accuracy, it generates verbose reasoning.
In contrast, MFT achieves higher accuracy at 45.7\% while consuming only 26.4\% of full CoT's reasoning overhead.
This confirms that MSC is not merely compressed reasoning but a more effective form that filters noise and streamlines logical flow, enabling better performance with significantly reduced computational cost.

{\normalsize $\blacktriangleright$ } \textit{Problem-Adaptive Threshold.}
Static thresholds $\delta \in [0.5, 0.9]$ exhibit clear accuracy-efficiency trade-offs.
High thresholds retain excessive reasoning, while low thresholds sacrifice critical reasoning steps, leading to noticeable performance degradation.
Among static settings, $\delta=0.7$ achieves the best balance.
Nevertheless, problem-adaptive thresholds naturally align with problem demands, surpassing all static configurations with comparable token usage.

{\normalsize $\blacktriangleright$ } \textit{Percentile-Based Complexity Estimation.}
We compare against two alternatives: 
Min-Max estimation $\mathcal{C}(x_i) = \frac{(\|z_i\| - \min_j \|z_j\|)}{(\max_j \|z_j\| - \min_j \|z_j\|)}$ 
and Log-Scaled normalization $\mathcal{C}(x_i) = \frac{\log(1+\|z_i\|) - \log(1+\min_j \|z_j\|)}{\log(1+\max_j \|z_j\|) - \log(1+\min_j \|z_j\|)}$.
Min-Max estimation is highly sensitive to outliers,
a single extremely long trajectory compresses all other samples into a narrow range, resulting in poor complexity discrimination.
Log-Scaled normalization partially alleviates this issue but still results in skewed scaling.
In contrast, percentile-based method produces a uniform complexity distribution, ensuring stable threshold scaling across diverse problems.

{\normalsize $\blacktriangleright$ } \textit{MSC Refinement.}
Without refinement, directly truncated CoT trajectories often results in abrupt or incomplete logical transitions.
The refinement process bridges these logical gaps while simultaneously eliminating redundancy, producing more coherent and concise reasoning chains.
Consequently, refinement reduces reasoning length by 38.6\% and boosts accuracy by 5.1\%.
Prompts along with concrete examples are provided in Appendix~\ref{sec:msc_refinement}.

\begin{figure}[ht]
\vskip 0.2in
\begin{center}
\centerline{\includegraphics[width=\columnwidth]{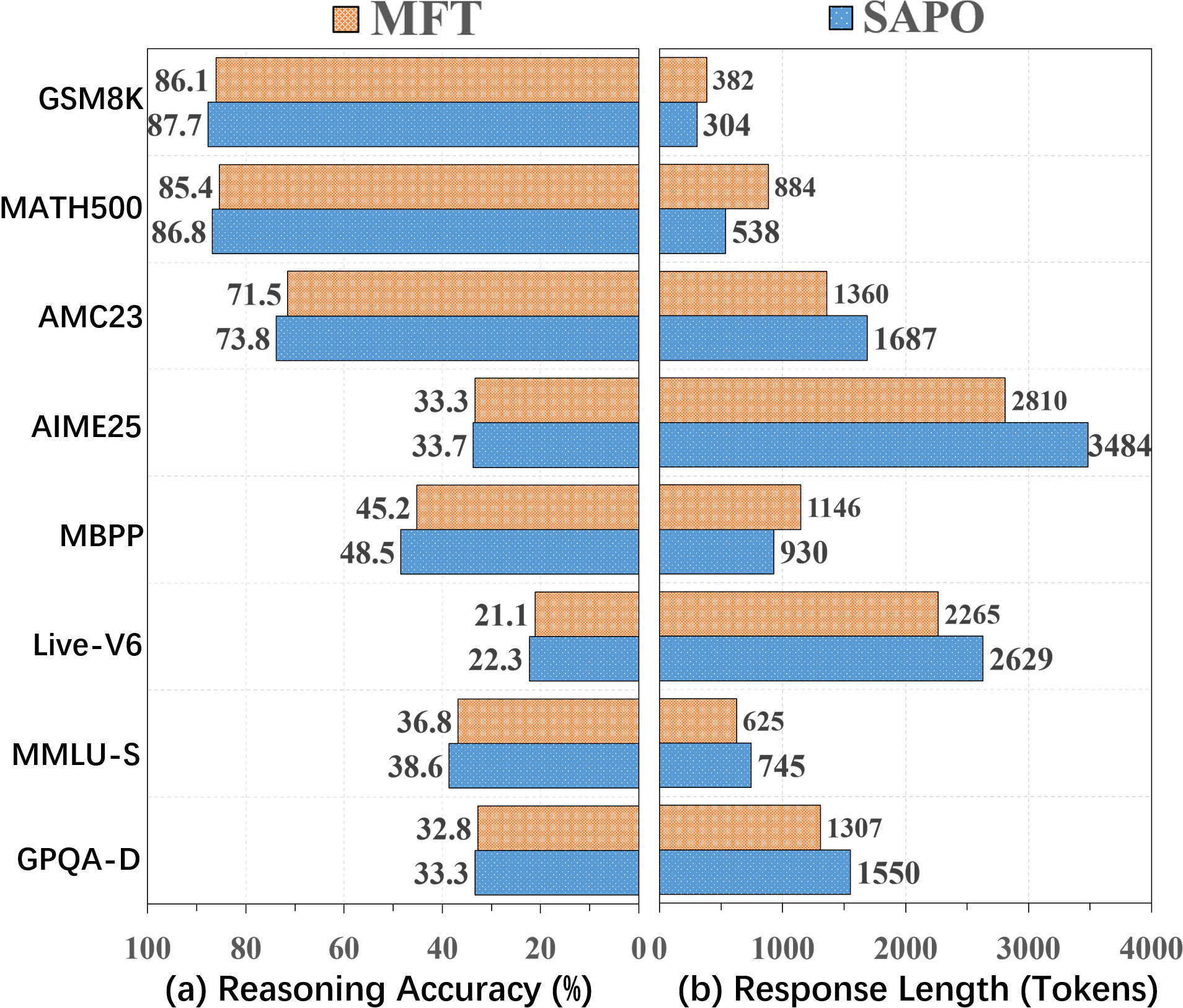}}
\caption{Per-benchmark accuracy and response length comparison. SAPO adaptively reduces reasoning on easier benchmarks while allocating more resources to challenging ones.}
\label{fig:sapo_comparison}
\end{center}
\vskip -0.2in
\end{figure}

\begin{table}[t]
\centering
\small
\caption{Ablation study of SAPO components. Dynamic Complexity Pool (DCP) and Sufficiency-Aware Reward Shaping($R_{suff}$).}
\label{tab:sapo_ablation}
\begin{tabular}{lcc}
\toprule
\textbf{Method} & \textbf{Accuracy (\%) $\uparrow$} & \textbf{Response Tokens $\downarrow$} \\
\midrule
MFT & 51.5 & 1,347 \\
\rowcolor{table_color}
\textbf{SAPO} & 53.1 & 1,483 \\
\midrule
w/o DCP & 52.9 & 1,642 \\
w/o $R_{suff}$ & 52.7 & 2,053 \\
\bottomrule
\end{tabular}
\end{table}
\paragraph{SAPO Ablations.}
We summarize ablation results in Table~\ref{tab:sapo_ablation} and visualize the per-benchmark accuracy-efficiency trade-offs in Figure~\ref{fig:sapo_comparison}.

{\normalsize $\blacktriangleright$ } \textit{Overall Effectiveness.}
Although SAPO slightly increases the response length, 
it improves accuracy across all benchmarks.
Crucially, this response increase does not reflect redundant reasoning.
As shown in Figure~\ref{fig:sapo_comparison}, 
on simple benchmarks where MFT already achieves high accuracy, SAPO successfully reduces reasoning.
Conversely, on challenging benchmarks, SAPO intelligently allocates additional reasoning budget.
This behavior indicates that SAPO learns to calibrate reasoning effort based on problem demands.

{\normalsize $\blacktriangleright$ } \textit{Dynamic Complexity Pool.}
In this ablation(w/o DCP), the complexity pool is initialized using MFT predictions but remains fixed during RL training.
Without online EMA updates, the estimated complexity gradually drifts away from the evolving policy.
This misalignment results in stale thresholds that fail to provide accurate sufficiency targets.

{\normalsize $\blacktriangleright$ } \textit{Sufficiency-Aware Reward Shaping.}
When the sufficiency reward is removed (w/o $R_{\text{suff}}$),
SAPO degenerates to vanilla GRPO that optimizes only correctness and format,
collapsing to verbose, full-CoT-style reasoning patterns.
The sufficiency reward provides fine-grained feedback on both over-thinking and under-thinking, encouraging concise yet reliable reasoning.

\subsection{Analysis}

\paragraph{Difficulty-conditioned reasoning length.}
\begin{figure}[ht]
\vskip 0.2in
\begin{center}
\centerline{\includegraphics[width=\columnwidth]{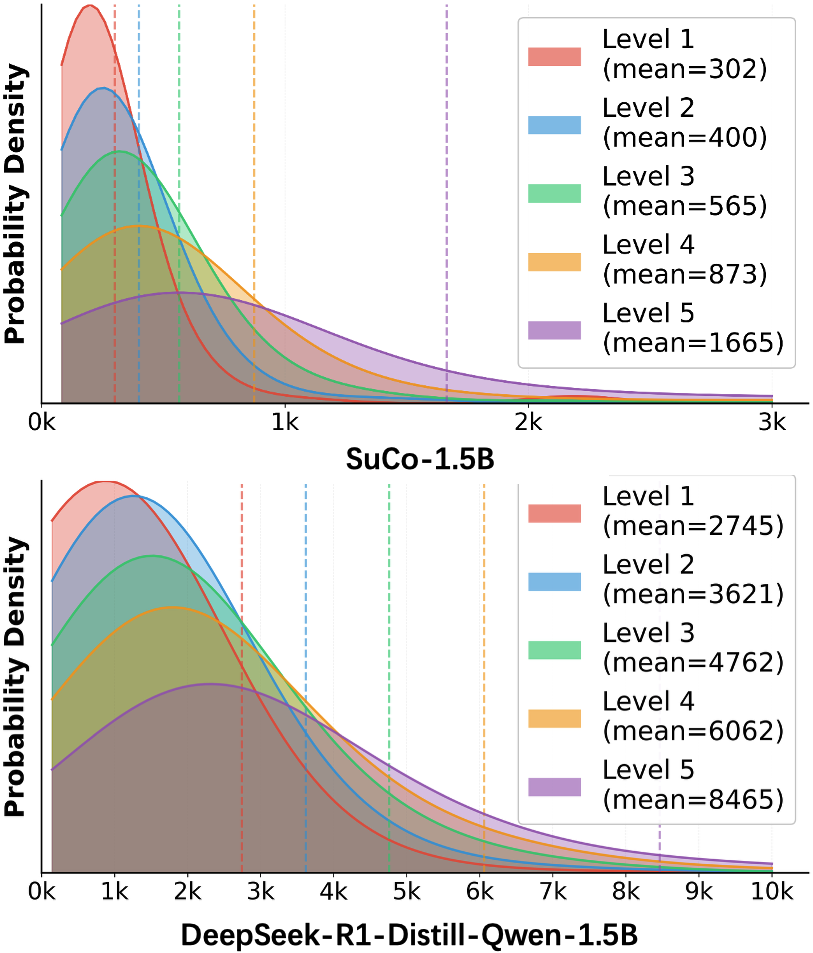}}
\caption{Response length distribution across MATH difficulty levels for SuCo-1.5B (top) and base LRM DeepSeek-R1-Distill-1.5B (bottom). 
SuCo continuously adapts reasoning effort to problem complexity with significantly higher efficiency.}
\label{fig:token_density_by_level}
\end{center}
\vskip -0.2in
\end{figure}

We compare response length distributions across MATH~\citep{math} difficulty levels between SuCo-1.5B and DeepSeek-R1-Distill-1.5B.
As shown in Figure~\ref{fig:token_density_by_level},
both models shift rightward as difficulty increases, but SuCo exhibits a much higher difficulty-sensitivity ratio:
the Level~5/Level~1 mean token ratio is $\approx 5.5\times$ for SuCo versus $\approx 3.1\times$ for the base LRM, indicating more discriminative resource allocation.

Moreover, SuCo operates in a fundamentally more efficient regime. On Level~1 problems, it uses 89\% fewer tokens than the base LRM while maintaining accuracy.
The base LRM's length variation reflects an inability to truncate unnecessary reasoning even for trivial queries, whereas SuCo's variation reflects genuine difficulty-conditioned allocation learned through sufficiency-aware training.

\paragraph{Out-of-Domain Generalization.}
To assess whether SuCo's adaptive reasoning capability generalizes beyond the training domains,
we conduct out-of-domain (OOD) evaluations on StrategyQA~\citep{geva2021strategyqa}, CommonsenseQA~\citep{talmor2019commonsenseqa}, and AlpacaEval 2.0~\citep{alpaca_eval}.
These tasks differ from the training distribution.

\begin{table}[ht]
\centering
\small
\caption{Out-of-domain generalization results. SuCo demonstrates strong transfer of adaptive reasoning to unseen task types.}
\label{tab:ood}
\setlength{\tabcolsep}{4pt}
\begin{tabular}{@{}l@{\hspace{6pt}}ccc@{}}
\toprule
\textbf{Method} & \textbf{StrategyQA} & \textbf{CSQA} & \textbf{AlpacaEval} \\
& ACC / Tok & ACC / Tok & LC-WR / Tok \\
\midrule
\makecell[l]{DeepSeek-R1\\-Distill} & 53.3 / 483 & 45.0 / 743 & 1.05 / 596 \\
Full CoT SFT & 22.6 / 742 & 19.4 / 1,061 & 0.3 / 743 \\
MFT & 28.0 / 213 & 26.6 / 342 & 0.67 / 314 \\
\rowcolor{table_color}
\textbf{SuCo} & \textbf{55.7} / 442 & \textbf{49.3} / 369 & \textbf{2.4} / 288 \\
\bottomrule
\end{tabular}
\end{table}

As shown in Table~\ref{tab:ood}, SuCo substantially outperforms all baselines on OOD tasks.
Notably, while MFT alone overfits to training domain patterns and degrades on OOD tasks, the SAPO stage enables SuCo to learn a generalizable policy for calibrating reasoning effort.

\paragraph{Cross-Model Robustness of MSC.}
To verify that MSC boundaries are robust across model families,
we construct MSC data using different calibrator models (Qwen3-4B, Qwen3-14B, DeepSeek-R1-Distill-Qwen-7B) and train different target models.
As shown in Table~\ref{tab:cross_model}, 
MSC supervision from all calibrators consistently outperforms full-CoT training across target models,
confirming that the constructed datasets transfer well across model families.

\begin{table}[ht]
\centering
\small
\caption{Cross-model robustness. Different calibrator models produce MSC data that consistently improves over Full CoT SFT across target model families. Format: Accuracy (Tokens).}
\label{tab:cross_model}
\begin{tabular}{lcc}
\toprule
\textbf{Calibrator} & \textbf{Qwen2.5-1.5B} & \textbf{Llama-3.2-3B} \\
\midrule
Full CoT SFT & 37.4 (5,177) & 38.1 (5,084) \\
\midrule
Qwen3-4B & 44.7 (1,394) & 44.2 (1,524) \\
Qwen3-14B & 44.2 (1,521) & 43.2 (1,821) \\
DS-R1-Distill-7B & 44.5 (1,491) & 43.7 (1,691) \\
\bottomrule
\end{tabular}
\end{table}

\paragraph{Empty CoT Analysis.}
\begin{figure}[t]
\begin{center}
\centerline{\includegraphics[width=\columnwidth]{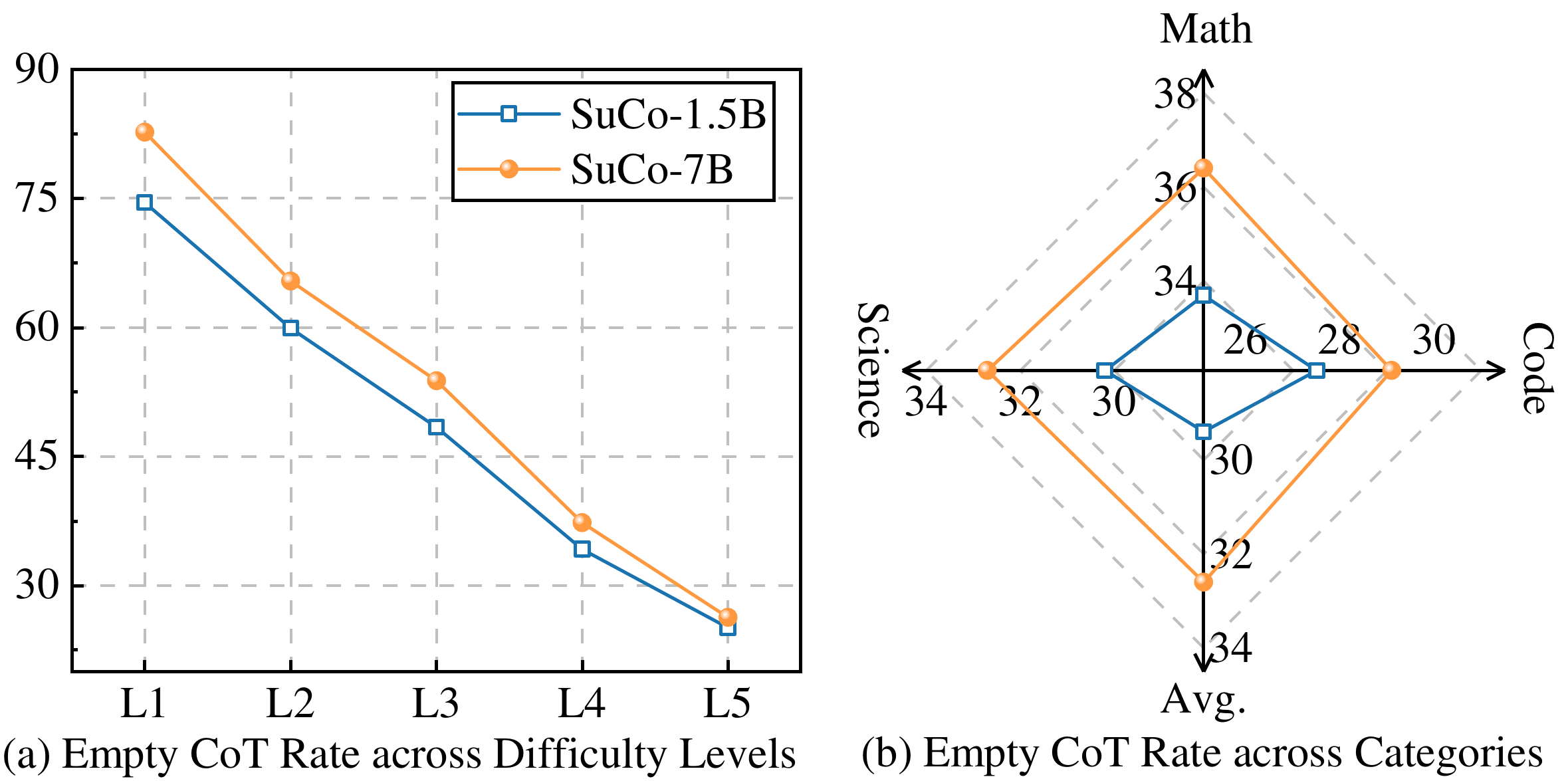}}
\caption{
Empty CoT analysis of SuCo-1.5B and SuCo-7B across problem types and difficulties. Higher model capacity (7B vs. 1.5B) leads to increased empty CoT rates,
while harder problems trigger more explicit reasoning.}
\label{fig:empty_cot}
\end{center}
\vskip -0.2in
\end{figure}

SuCo learns to skip explicit reasoning when problems are trivial,
directly outputting answers without explicit reasoning.
Figure~\ref{fig:empty_cot}(a) reveals that empty CoT rates decrease monotonically with increasing difficulty,
indicating that the model increasingly engages in explicit reasoning for harder problems.
The 7B model consistently exhibits higher empty rates than 1.5B across all levels,
reflecting its stronger capabilities that reduce reliance on intermediate reasoning steps.

Across domains (Figure~\ref{fig:empty_cot}(b)),
empty CoT rates remain relatively stable at ~30--37\%,
suggesting that the decision to omit explicit reasoning is largely task-agnostic.
Math problems show a slightly higher proportion of empty responses,
likely due to formula-based questions requiring minimal explicit derivation.
Despite a substantial fraction of empty CoT outputs, SuCo maintains strong overall accuracy
(Table~\ref{tab:main_results}), 
suggesting that explicit reasoning is not always necessary,
and that selectively omitting CoT can preserve or even improve efficiency without sacrificing accuracy.



\section{Conclusion}
In this work, we formalize \emph{Minimal Sufficient CoT} (MSC) as the shortest reasoning prefix adequate for correct answers, revealing that models can perform better with less reasoning. 
Building on this insight, we propose \emph{Sufficiency-guided Continuous Adaptive Reasoning} (SuCo), a two-stage framework enabling continuous and autonomous reasoning adaptation.
Through \emph{MSC-Aligned Fine-Tuning} (MFT) and \emph{Sufficiency-Aware Policy Optimization} (SAPO), SuCo learns to calibrate its reasoning effort according to problem demands without relying on discrete modes or external controllers.
Extensive experiments across mathematics, code, and science benchmarks demonstrate that SuCo consistently achieves higher accuracy with significantly fewer reasoning tokens.

\paragraph{Limitations.}
We acknowledge several limitations.
First, MSC construction relies on ground-truth answers to compute sufficiency scores, 
which limits direct application to open-ended generation tasks.
However, once trained, the model internalizes adaptive reasoning as a general capability.
Second, the MFT stage depends on data distilled from strong LRMs.
While removing the 80B refinement model still yields results superior to all baselines, 
reducing this dependency remains desirable.

\paragraph{Future Work.}
Extending sufficiency estimation to open-ended settings is a promising avenue.
Additionally, agentic tasks present a compelling application scenario,
where over-thinking incurs redundant API costs and under-thinking leads to task failure.
Extending SuCo to such settings is a promising direction.
\section*{Impact Statement}

This work aims to advance the field of machine learning by proposing a more efficient and adaptive training framework for reasoning models.
Our method focuses on technical efficiency improvements and does not alter the fundamental capabilities or safety properties of underlying models.
We do not foresee any ethical concerns or societal consequences beyond those commonly associated with research on large language models.

\section*{Acknowledgements}
This work was supported in part by National Natural Science Foundation of China (62476070), Shenzhen Science and Technology Program (JCYJ20241202123503005,  GXWD20231128103232001, ZDSYS20230626091203008, KQTD20240729102154066), Department of Science and Technology of Guangdong (2024A1515011540) and National Key R\&D Program of China (SQ2024YFE0200592).

\bibliography{references} 
\bibliographystyle{icml2026}

\newpage
\appendix
\onecolumn

\section{Additional Ablation Studies}
\label{app:additional_ablations}

\subsection{Minimum Reasoning Threshold $L_{\min}$}
\label{app:lmin}
During MSC construction, if the raw MSC prefix contains fewer than or equal to $L_{\min}$ sentences, we set it to an empty string, indicating the model should directly generate the answer without intermediate reasoning steps.

\begin{figure*}[ht]
\centering
\includegraphics[width=\textwidth]{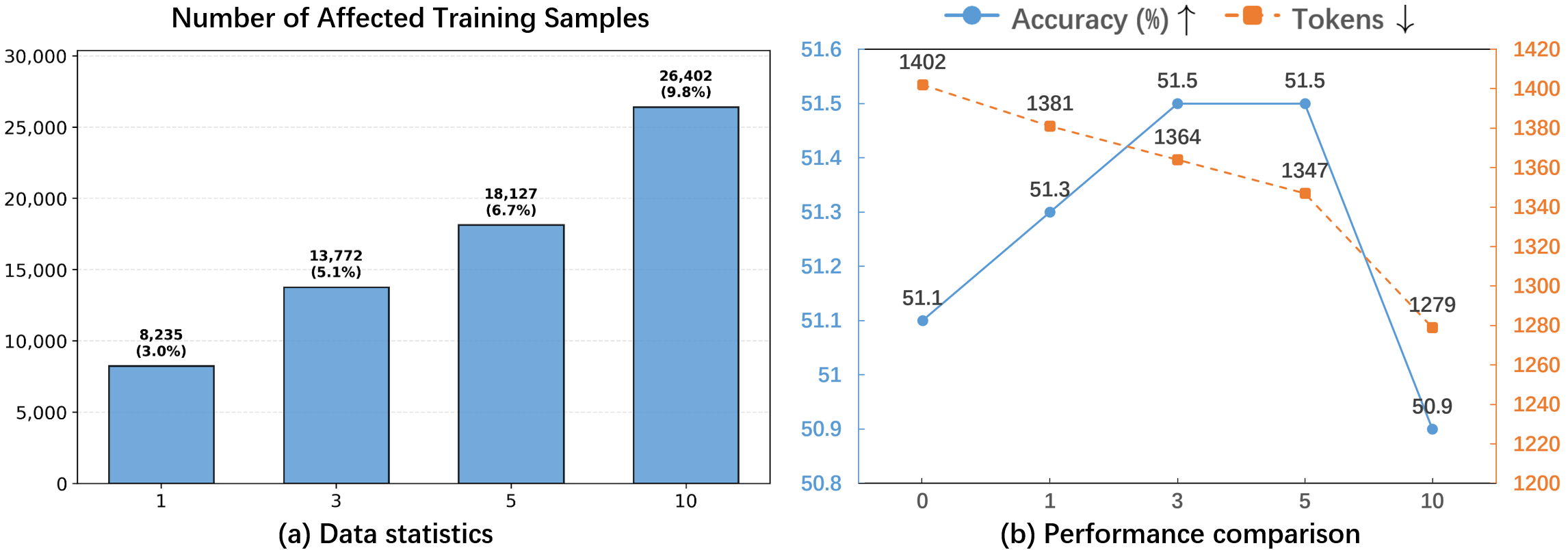}
\caption{Effect of the minimum reasoning length $L_{\min}$.}
\label{fig:lmin_ablation}
\end{figure*}

Figure~\ref{fig:lmin_ablation}(a) shows the number of affected training samples at different thresholds. As $L_{\min}$ increases from 1 to 10, the proportion of non-thinking samples grows from 3.0\% to 9.8\%.

As shown in Figure~\ref{fig:lmin_ablation}(b), 
without the threshold ($L_{\min}=0$), trivial CoT fragments introduce noise, resulting in 51.1\% accuracy with 1,402 tokens. Overly aggressive filtering ($L_{\min}=10$) suppresses necessary reasoning, degrading accuracy to 50.9\%. $L_{\min}=5$ achieves the optimal balance at 51.5\% accuracy with 1,347 tokens, demonstrating that filtering very short CoT fragments (affecting 6.7\% of samples) effectively removes noise while preserving meaningful reasoning signals.

\subsection{EMA Rate $\eta$.}
\label{app:eta}
\begin{wraptable}{r}{0.4\textwidth}
\centering
\small
\caption{Effect of EMA rate $\eta$.}
\label{tab:eta_ablation}
\begin{tabular}{ccc}
\toprule
\textbf{EMA Rate ($\eta$)} & \textbf{Accuracy $\uparrow$} & \textbf{Tokens $\downarrow$} \\ \midrule 0.0 (w/o DCP) & \underline{52.9} & 1,642 \\ \rowcolor{table_color} \textbf{0.1} & \textbf{53.1} & \textbf{1,483} \\ 0.3 & \underline{52.9} & 1,442 \\ 0.5 & 52.6 & 1,427 \\ 1.0 (Full Update) & 52.1 & \underline{1,369} \\
\bottomrule
\end{tabular}
\end{wraptable}
We analyze the impact of EMA rate $\eta$ on the dynamic complexity pool update.
As shown in Table~\ref{tab:eta_ablation}, $\eta=0.1$ achieves the best accuracy-efficiency balance at 53.1\% accuracy with 1,483 tokens.
Static pool ($\eta=0$) retains more redundant reasoning (1,642 tokens) while achieving comparable accuracy (52.9\%).
Overly aggressive updates ($\eta \geq 0.5$) reduce token usage but degrade accuracy due to unstable threshold estimation.
This validates that moderate EMA rates effectively balance tracking policy evolution with maintaining stable training signals.

\subsection{Over-thinking Tolerance $\epsilon$}
\label{app:epsilon}

The tolerance parameter $\epsilon$ in Eq.~\ref{eq:suff_reward} controls the strictness of over-thinking penalties by allowing minor deviations beyond the minimal sufficient prefix.

\begin{wraptable}{r}{0.4\textwidth} 
\centering
\small
\caption{Effect of over-thinking tolerance $\epsilon$ on SAPO performance. Results are averaged across all benchmarks on Qwen2.5-Math-1.5B.}
\label{tab:epsilon_ablation}
\begin{tabular}{ccc}
\toprule
\textbf{Tolerance ($\epsilon$)} & \textbf{Accuracy (\%) $\uparrow$} & \textbf{Tokens $\downarrow$} \\
\midrule
0 (Strict) & 52.4 & 1,391 \\
1 & 52.8 & 1,456 \\
\rowcolor{table_color}
\textbf{2} & \textbf{53.1} & \textbf{1,483} \\
3 & 53.0 & 1,527 \\
5 & 52.7 & 1,658 \\
\bottomrule
\end{tabular}
\end{wraptable}

As shown in Table~\ref{tab:epsilon_ablation}, setting $\epsilon=0$ applies strict penalties for any reasoning beyond the minimal sufficient prefix, resulting in overly aggressive truncation that reduces tokens to 1,391 but harms accuracy (52.4\%).
This strict constraint prevents the model from generating natural reasoning flow and exploring slightly longer but potentially more robust reasoning paths.

With moderate tolerance ($\epsilon=2$), the model achieves the best accuracy at 53.1\% while generating 1,483 tokens.
This tolerance allows the model to extend reasoning by 1-2 sentences beyond the minimal sufficient point when beneficial, accommodating natural variations in reasoning style without sacrificing efficiency.

As $\epsilon$ increases further (3, 5), accuracy plateaus or slightly declines while token usage grows substantially.
At $\epsilon=5$, the sufficiency constraint becomes too loose, allowing the model to generate verbose reasoning (1,658 tokens) that approaches the behavior without sufficiency-aware rewards.
This demonstrates that $\epsilon=2$ provides an appropriate balance: it avoids overly rigid constraints that harm reasoning quality while maintaining effective control over redundant thinking.

\subsection{Sufficiency Metric Ablation}
\label{app:sufficiency_ablation}

We compare our geometric mean sufficiency formulation against alternative definitions 
to justify the design choice.
All variants use the same MSC construction pipeline with Qwen2.5-Math-1.5B as the target model.

\begin{table}[ht]
\centering
\small
\caption{Comparison of sufficiency metric formulations. Geometric mean provides the best balance between accuracy and efficiency.}
\label{tab:sufficiency_metric}
\begin{tabular}{lccc}
\toprule
\textbf{Sufficiency Metric} & \textbf{Train CoT Tokens} & \textbf{Accuracy (\%)} & \textbf{Inference Tokens} \\
\midrule
Full CoT SFT (no truncation) & 3,781 & 38.4 & 5,082 \\
\midrule
Joint Probability & 2,149 & 40.1 & 2,573 \\
Arithmetic Mean & 1,341 & 43.1 & 1,578 \\
\rowcolor{table_color}
\textbf{Geometric Mean (Ours)} & \textbf{1,138} & \textbf{45.7} & \textbf{1,344} \\
\bottomrule
\end{tabular}
\end{table}

\textbf{Joint Probability} ($\prod_i \pi_\theta(y^*_i | \cdot)$) decays exponentially with answer length, 
causing the threshold to be satisfied too late for short-answer problems and too early for long-answer problems.
This results in inconsistent truncation quality.

\textbf{Arithmetic Mean} ($\frac{1}{\|y^*\|}\sum_i \pi_\theta(y^*_i | \cdot)$) is dominated by a few high-confidence tokens, 
making it less sensitive to tokens that genuinely require reasoning support.

\textbf{Geometric Mean} (Eq.~\ref{eq:sufficiency}) normalizes joint probability into per-token average log-probability,
which is stable across varying answer lengths and equally sensitive to all answer tokens.
It achieves the highest accuracy with the most aggressive token reduction,
confirming its effectiveness as a sufficiency signal.

\subsection{Cross-Domain vs.\ Intra-Domain Percentile}
\label{app:cross_domain}

In our default setting, complexity percentiles are computed globally across all training domains.
However, different domains exhibit different baseline reasoning lengths. For instance, code problems typically require longer traces than math problems.
This raises the question of whether a code problem might be assigned an artificially high complexity score simply because code traces are longer on average, rather than because the problem itself is harder.
To investigate this, we compare the default cross-domain percentile with an intra-domain variant that computes percentiles separately within each domain (math, code, science).

\begin{table}[ht]
\centering
\small
\caption{Cross-domain vs.\ intra-domain percentile estimation (Qwen2.5-Math-1.5B MFT).}
\label{tab:cross_domain}
\begin{tabular}{l|cc|cc|cc|cc}
\toprule
& \multicolumn{2}{c|}{\textbf{Math}} & \multicolumn{2}{c|}{\textbf{Code}} & \multicolumn{2}{c|}{\textbf{Science}} & \multicolumn{2}{c}{\textbf{Avg.}} \\
\textbf{Method} & Acc & Tok & Acc & Tok & Acc & Tok & Acc & Tok \\
\midrule
Full CoT SFT & 59.1 & 5,380 & 28.5 & 6,345 & 27.7 & 3,522 & 38.4 & 5,082 \\
Cross-domain & 69.1 & 1,359 & 33.2 & 1,706 & 34.8 & 966 & 45.7 & 1,344 \\
Intra-domain & 69.2 & 1,376 & 33.2 & 1,692 & 35.1 & 1,021 & 45.8 & 1,363 \\
\bottomrule
\end{tabular}
\end{table}

Intra-domain percentile yields nearly identical performance to the cross-domain setting,
indicating that the global percentile preserves monotonicity within each domain 
and remains a robust measure of reasoning difficulty.
This robustness arises because percentile ranks maintain relative ordering within domains,
regardless of absolute length differences across domains.
\section{MSC Dataset Construction}
\label{sec:dataset_construction}

\subsection{Dataset Statistics}
Table~\ref{tab:dataset_stats} summarizes the statistics of the final dataset.
Across all samples, the full CoT traces average 3,781 tokens, while MSC reduces this to 1,138 tokens.
Notably, 109,882 samples (40.7\%) yield empty MSCs, indicating that the model can solve these problems without explicit reasoning.

For Stage \RN{1}, we train on all samples to learn minimal sufficient reasoning patterns. 
For Stage \RN{2}, we sample 50,000 instances for RL to balance training efficiency and diversity.

\begin{table}[ht]
\caption{Training dataset statistics. We report the number of samples, average token counts for full CoT and MSC, and the number of samples with empty MSC (requiring no explicit reasoning).}
\label{tab:dataset_stats}
\vskip 0.15in
\begin{center}
\begin{small}
\begin{sc}
\begin{tabular}{c l r r r r r}
\toprule
Domain & Source & Samples & Full CoT & Raw MSC & Refine MSC & Empty \\
\midrule
\multirow{4}{*}{Math}     & Llama-Nemotron         & 39,377 & 2,924 & 1,238 & 1,067 & 10,683 \\
                          & Mixture-of-Thoughts    & 51,089 & 5,414 & 2,195 & 1,424 & 14,684 \\
                          & OpenR1-Math-220k       & 37,899 & 4,482 & 1,934 & 1,217 & 13,997 \\  
                          & s1K-1.1                & 330    & 7,996 & 3,410 & 1,681 & 75     \\  
\midrule
\multirow{3}{*}{Code}     & Llama-Nemotron         & 60,000 & 2,263 & 1,329 & 537   & 24,837 \\
                          & Mixture-of-Thoughts    & 12,671 & 11,106 & 2,764 & 2,379 & 2,914  \\
                          & OpenCodeReasoning      & 15,704 & 6,076 & 2,491 & 1,599 & 3,424  \\
\midrule
\multirow{2}{*}{Science}  & Mixture-of-Thoughts    & 52,876 & 1,590 & 1,597  & 623   & 39,242  \\  
                          & s1K-1.1                & 65     & 8,644 & 1,179  & 2,178 & 26 \\  
\midrule
\multicolumn{2}{l}{Total}     & 270,011 & --    & -- & --    & 109,882   \\
\multicolumn{2}{l}{Average}   & --      & 3,781 & 1,803 & 1,138    &  --      \\
\bottomrule
\end{tabular}
\end{sc}
\end{small}
\end{center}
\vskip -0.1in
\end{table}

\subsection{Data Sources}
We curate our training data from the following five publicly available reasoning datasets,
all containing CoT trajectories distilled from advanced LRMs:
\begin{itemize}
    \item \textbf{Mixture-of-Thoughts}~\citep{mot}: 350K samples (93K math, 83K code, 173K science) generated by DeepSeek-R1 with correctness filtering on final answers.
    \item \textbf{OpenR1-Math-220k}~\citep{openr1math}: 220K math reasoning trajectories distilled from 800K DeepSeek-R1 generated solutions.
    \item \textbf{Llama-Nemotron Post-Training Dataset}~\citep{llama}: 3.9M samples covering math, code, science, chat, and safety. All samples include explicit reasoning trajectories produced by DeepSeek-R1 and refined using Nemotron-340B~\citep{adler2024nemotron}.
    \item \textbf{OpenCodeReasoning}~\citep{opencodereasoning}: 735K Codeforces/LeetCode problems paired with CoT and executable Python solutions, including full test cases.
    \item \textbf{s1K-1.1}~\citep{s1k}: 1,000 carefully curated high-difficulty examples selected for difficulty, diversity, and quality, with accompanying budget-constrained inference technique.
\end{itemize}

\subsection{Data Preprocessing}
We apply a rigorous preprocessing pipeline to ensure data quality:

\textbf{Filtering.} 
We remove samples with: 
(1) incorrect or missing answers, 
(2) incomplete reasoning traces, 
(3) overlap with our evaluation benchmarks, and 
(4) embedded non-textual elements (e.g., images, URLs),
(5) non-English content.

\textbf{Deduplication.} 
We apply MinHash LSH~\citep{lee2022deduplicating} to remove near-duplicate samples.

\textbf{Cleaning.} 
Questions are normalized by removing source identifiers and numbering to reduce stylistic noise.

\subsection{MSC Construction}
For each sample, we derive its MSC following Algorithm~\ref{alg:msc}. 
Additionally, we employ an LLM-based evaluation to score each MSC along three dimensions:
(1) correctness,
(2) sufficiency and support for the final answer,
(3) fluency and logical coherence.
Low-quality MSC samples are filtered out. 
Both MSC refinement and quality assessment are performed  with Qwen3-Next-80B-A3B-Instruct~\citep{qwen3}.

\section{MSC Refinement}
\label{sec:msc_refinement}

\subsection{Refinement Prompt}
\begin{figure*}[ht]
\centering
\includegraphics[width=\textwidth]{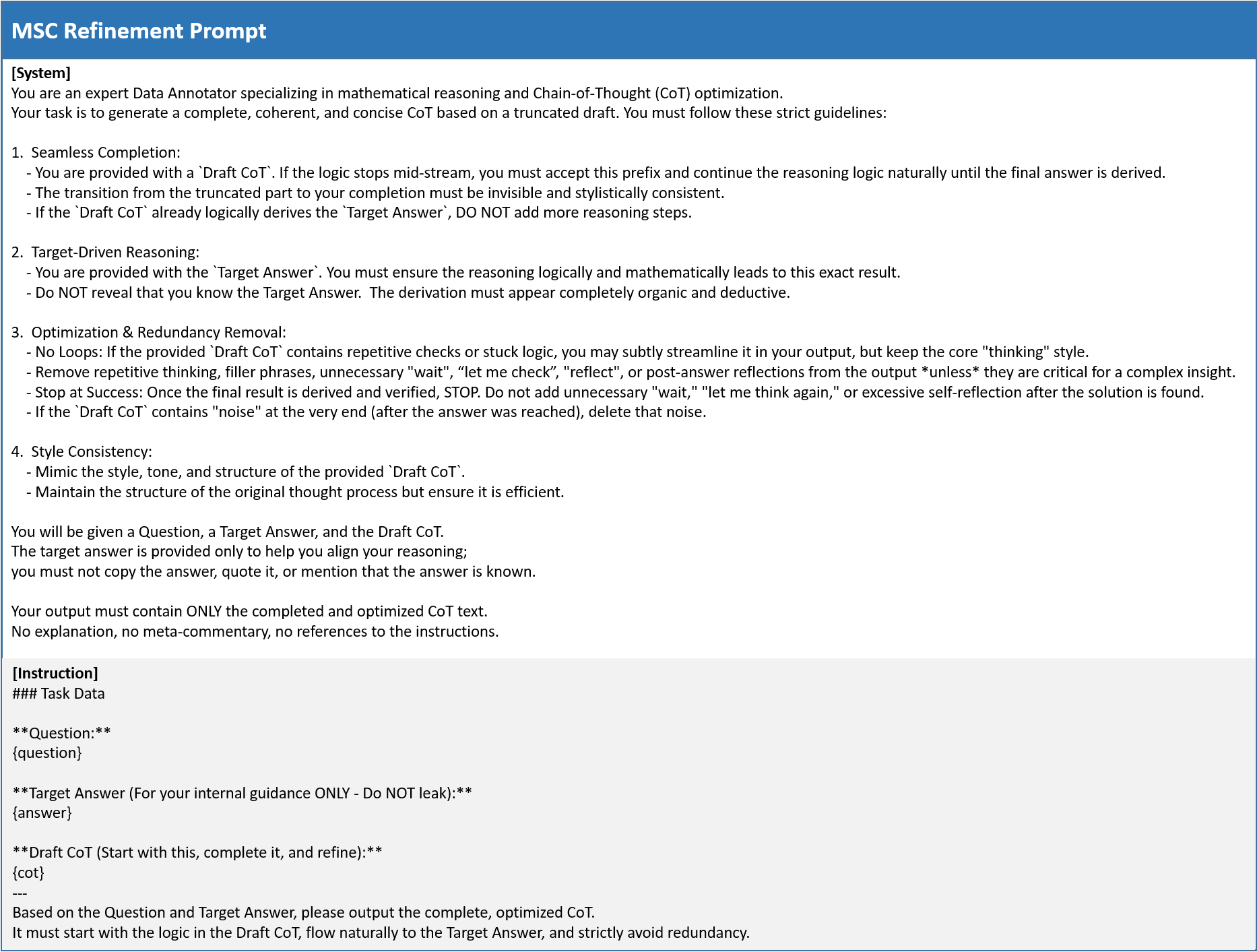}
\caption{Complete prompt for MSC refinement. The prompt guides the refinement model to enhance logical completeness and conciseness while maintaining stylistic consistency with the original reasoning trajectory.}
\label{fig:msc_prompt}
\end{figure*}

Figure~\ref{fig:msc_prompt} presents the complete prompt used for MSC refinement.
The prompt guides the model to polish the raw MSC prefix along three dimensions:
\textbf{Logical Completeness}, \textbf{Conciseness}, and \textbf{Stylistic Consistency}.
The refinement process focuses on improving coherence and readability of the existing MSC
without modifying its underlying reasoning content.

\subsection{Refinement Examples}

\begin{figure*}[ht]
\centering
\includegraphics[width=\textwidth]{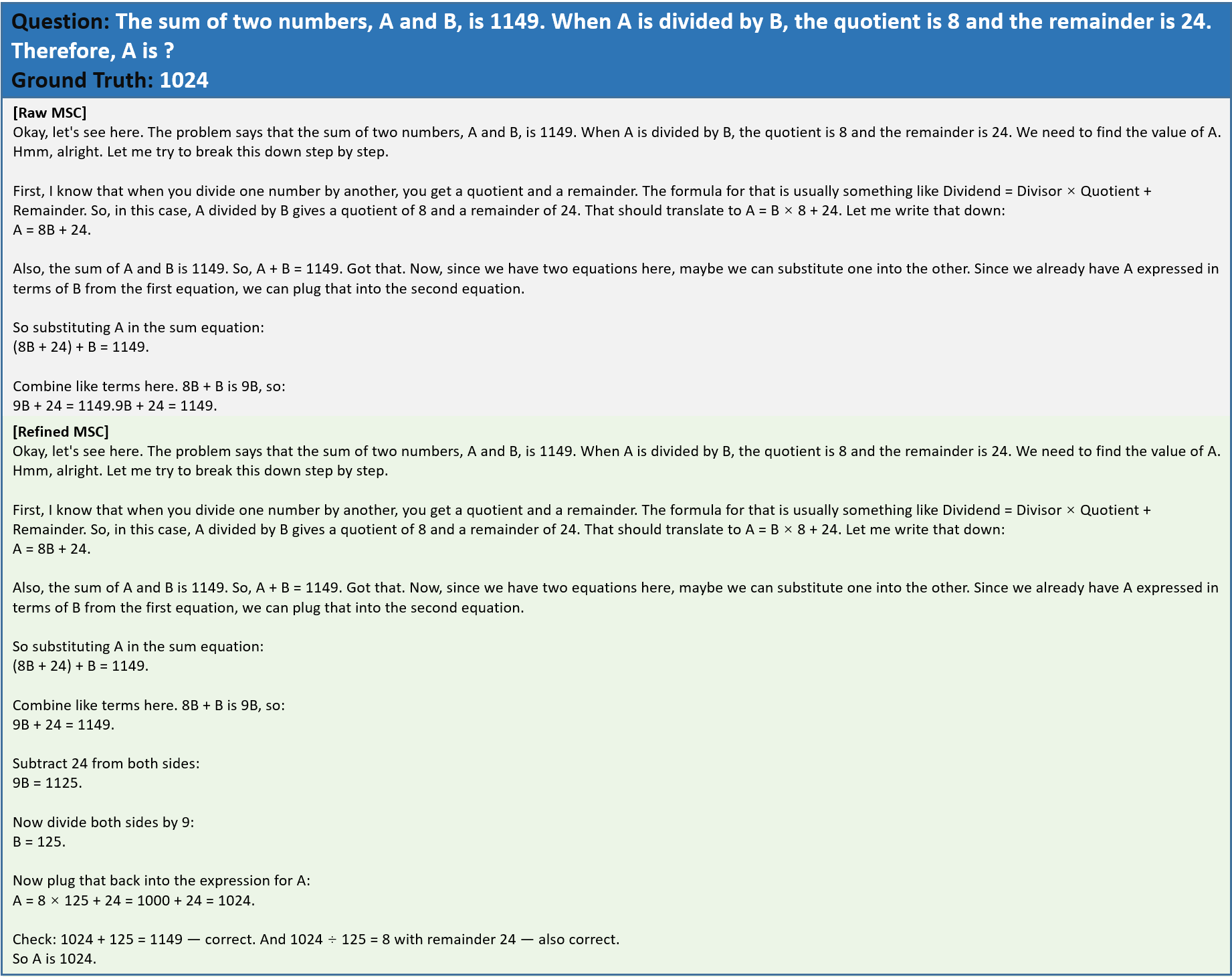}
\caption{Refinement example demonstrating logical completion. Raw MSC stops mid-reasoning; refined MSC completes the derivation while preserving the original flow.}
\label{fig:msc_refine_case1}
\end{figure*}

\begin{figure*}[ht]
\centering
\includegraphics[width=\textwidth]{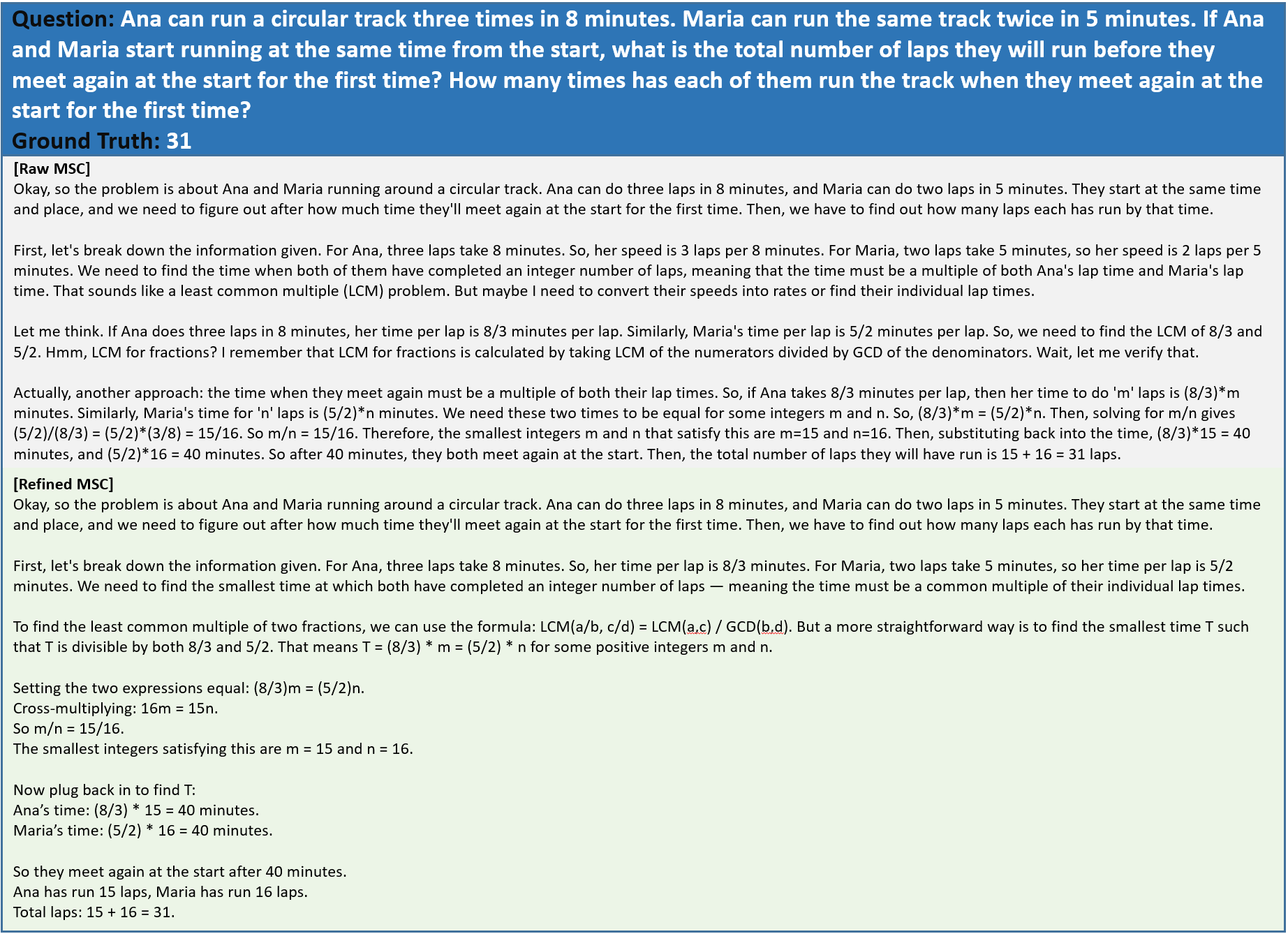}
\caption{Refinement example: reasoning optimization. Raw MSC contains exploratory backtracking; refined MSC eliminates redundancy while maintaining the core logic.}
\label{fig:msc_refine_case2}
\end{figure*}

Figures~\ref{fig:msc_refine_case1} and~\ref{fig:msc_refine_case2} illustrate concrete examples comparing raw MSC and refined MSC.

\end{document}